\documentclass{article}

\usepackage{arxiv}

\usepackage[utf8]{inputenc} 
\usepackage[T1]{fontenc}    
\usepackage{url}            
\usepackage[colorlinks=true, linkcolor=blue, urlcolor=cyan, citecolor=cyan]{hyperref}

\usepackage{booktabs}       
\usepackage{amsfonts}       
\usepackage{nicefrac}       
\usepackage{microtype}      
\usepackage{graphicx}
\usepackage{doi}
\usepackage{subcaption}
\usepackage{stfloats}
\usepackage{amsmath}
\usepackage{multirow}
\usepackage{color, colortbl}
\usepackage{float}
\usepackage{makecell}
\usepackage{caption}
\usepackage{setspace}
\usepackage{svg}
\usepackage{soul}
\usepackage{listings}
\usepackage{algorithm}
\usepackage{multicol}
\usepackage{appendix}

\title{Contrastive Forward-Forward: A Training Algorithm of Vision Transformer}

\definecolor{commentcolor}{RGB}{110,154,155}   

\author{Hossein Aghagolzadeh \\
	Faculty of Electrical and Computer Engineering \\
	  Babol Noshirvani University of Technology\\
	Babol, Iran \\
	\texttt{ho.golzadeh@stu.nit.ac.ir} \\
	\And
	Mehdi Ezoji \\
	Faculty of Electrical and Computer Engineering \\
	  Babol Noshirvani University of Technology\\
	Babol, Iran \\
	\texttt{m.ezoji@nit.ac.ir} \\
}

\date{}

\renewcommand{\shorttitle}{Contrastive Forward-Forward}

\hypersetup{
pdftitle={Contrastive Forward-Forward},
pdfsubject={},
pdfauthor={H.~Aghagolzadeh, M.~Ezoji},
pdfkeywords={Forward-Forward, Contrastive Learning, Image Classification, Vision Transformer, Backpropagation},
}

\begin{document}
\maketitle

\begin{abstract}
Although backpropagation is widely accepted as a training algorithm for artificial neural networks, researchers are always looking for inspiration from the brain to find ways with potentially better performance. Forward-Forward is a novel training algorithm that is more similar to what occurs in the brain, although there is a significant performance gap compared to backpropagation. In the Forward-Forward algorithm, the loss functions are placed after each layer, and the updating of a layer is done using two local forward passes and one local backward pass. Forward-Forward is in its early stages and has been designed and evaluated on simple multi-layer perceptron networks to solve image classification tasks. In this work, we have extended the use of this algorithm to a more complex and modern network, namely the Vision Transformer. Inspired by insights from contrastive learning, we have attempted to revise this algorithm, leading to the introduction of Contrastive Forward-Forward. Experimental results show that our proposed algorithm performs significantly better than the baseline Forward-Forward leading to an increase of up to 10\% in accuracy and accelerating the convergence speed by 5 to 20 times. Furthermore, if we take Cross Entropy as the baseline loss function in backpropagation, it will be demonstrated that the proposed modifications to the baseline Forward-Forward reduce its performance gap compared to backpropagation on Vision Transformer, and even outperforms it in certain conditions, such as inaccurate supervision\footnote{This is the preprint of our paper accepted in Neural Networks. 
The published version includes additional experiments (e.g., layer-wise parallel training) and is available at \href{https://doi.org/10.1016/j.neunet.2025.107867}{doi.org/10.1016/j.neunet.2025.107867}.}. Code is available at: \href{https://github.com/HosseinAghagol/ContrastiveFF}{github.com/HosseinAghagol/ContrastiveFF}.
\end{abstract}


\section{Introduction}
The brain is known to be a powerful cognitive tool, and researchers are always looking for inspiration from its structure and functions in various domains. In the machine learning, Artificial Neural Networks (ANNs) —the core of most solutions in this field— are a very simple model of the brain. However, it has been analyzed that the algorithm responsible for training those structures is very different from what happens in the brain \cite{lillicrap2020backpropagation}. This well-known algorithm is called backpropagation (BP) \cite{rumelhart1986learning}. Recently, a neural network training algorithm named Forward-Forward (FF) \cite{hinton2022forward} has appeared, claiming to be more similar to the processes that occur in the brain. Although BP has a very strong mathematical foundation, and imitation of biological functions is not always the best solution, a part of machine learning research consistently is looking for inspiration from biological mechanisms to improve methods. In the meantime, some challenges of using BP such as getting trapped in the local minima \cite{BALDI198953}, the vanishing/exploding gradient problem \cite{LU2023283} and overfitting \cite{lawrence2000overfitting} provide additional motivation to find alternative methods.

Classification is the most fundamental problem that ANNs face. Most algorithms and networks are initially designed and evaluated on classification tasks before being applied to other problems. Among these, image classification has been particularly popular due to its complexity and wide range of applications, so that the development of Convolutional Neural Networks (CNNs) \cite{lecun1998gradient} attracted immense attention \cite{krizhevsky2012imagenet, he2016deep}. Recently, transformer networks, which were initially used for Natural Language Processing (NLP) \cite{vaswani2017attention}, have been applied to image classification in the form of Vision Transformer (ViT) \cite{dosovitskiy2020image}, revitalizing the field of machine learning. Although there is a close competition between ViTs and CNNs in terms of performance in image-related tasks \cite{woo2023convnext}, ViTs have become more popular due to advantages such as better scalability for larger datasets \cite{dosovitskiy2020image, tolstikhin2021mlp} or ability to use similar model components across multiple modalities \cite{fuyu-8b}.

In summary, the BP algorithm for image classification is as follows: images are fed to the network and sequentially pass through different layers to produce the final output. These outputs are the predicted labels for the input images. These predicted labels are then compared to the true labels using a loss function. This loss is back-propagated from the end of the network using gradient descent-based methods, updating the network weights according to the computed gradients to improve future predictions \cite{rumelhart1986learning}. These two stages are known as \textit{forward} and \textit{backward}. During the \textit{backward}, to propagate gradients based on mathematical relationships, the outputs of all layers must be stored during the \textit{forward}, which requires sufficient memory for storage. In addition, subsequent batches of data cannot enter the network until \textit{forward} and \textit{backward} propagation have been completed \cite{hinton2022forward}. This operation of BP is different from what happens in the brain, making it implausible in the brain for two main reasons \cite{lillicrap2020backpropagation}: (i) In the brain, the data does not wait for the \textit{forward} and  \textit{backward} operations of the previous data. (ii) The storage of layer outputs and gradients has not been observed in the brain. FF was introduced focusing on these two points to increase the similarity between the brain function and the training algorithm.

FF proposes placing the loss function after each layer of the network instead of at the end. This means that the loss is calculated, and the layer is updated immediately after the output of each layer is ready, rather than waiting for the output of the entire network \cite{hinton2022forward}. This approach can be described as employing a local loss. The  challenge of this approach is that the predicted labels are not available in the intermediate layers; Therefore, using a loss function like Cross Entropy (CE), which directly compares the true label with the predicted label, would be meaningless as local loss. FF proposes to handle this challenge by using two forward passes: the first forward pass is done with the correct label, and the second forward pass is done with an incorrect label. In this way, by defining an appropriate loss function, the network can be trained after each layer to learn whether the image was entered with the correct label or not. If this is achieved, the layers become independent of each other, except for feeding their output to the next layer. FF can use parallelization between the network layers, significantly reducing memory requirements or substantially decreasing the training time per epoch. Additionally, since each layer has its own loss function, conceptually, the occurrence of the vanishing/exploding gradient problem becomes irrelevant.

FF calculates the local loss by comparing the representations obtained from positive samples, the samples with the correct labels, and negative samples, the same samples but with incorrect labels after each layer \cite{hinton2022forward}. From the point of view of comparing these two representations, FF is similar to another technique known as contrastive learning \cite{hadsell2006dimensionality, sohn2016improved, chen2020simple, radford2021learning}. Although contrastive learning is well known in its self-supervised form, there is a method called Supervised Contrastive Learning (SCL\footnote{Note that SCL is commonly used in the literature to refer to Self-supervised Contrastive Learning, but in this work, we refer to Supervised Contrastive Learning.}) \cite{khosla2020supervised} that extends it for use in supervised settings. In contrastive learning methods, the idea also includes two forward passes and a comparison of the resulting representations. This similarity inspired us to leverage the insights from contrastive learning to modify FF.

The key contributions of this work can be summarized as follows:

\begin{itemize}
    \item Inspired by SCL, the developed FF algorithm is presented as the proposed method regardless of the network architecture.
    
    \item With slight modifications to FF and the proposed method, they were extended for use in ViT without any changes to the network architecture.
    \item The contrastive loss has been modified to better align with its use in the proposed method.
    \item Through various experiments, it is demonstrated that the proposed method outperforms FF in terms of accuracy and convergence speed, also exhibiting only a small performance gap compared to BP.
    \item A practical pipeline for parallelizing the training of different layers of the network based on the proposed method is outlined.
\end{itemize}

The rest of this paper is organized as follows: In Sec. \ref{sec:related}, related studies in contrastive learning that inspired this work are discussed, and efforts to improve the FF are highlighted. In Sec. \ref{sec:method}, the fundamentals of SCL \cite{khosla2020supervised} and FF \cite{hinton2022forward} are presented, followed by the proposed method. In Sec. \ref{sec:exp}, the experiments and their results are described. In Sec. \ref{sec:discussion}, the parallelization and inductive bias in layer training discussed from the proposed method's perspective. In Sec. \ref{sec:conclusion}, the findings are summarized, the limitations of the proposed method are outlined, and future works are suggested.

\section{Related Work}\label{sec:related}
After the increased attention to deep learning \cite{krizhevsky2012imagenet}, the main focus of machine learning studies shifted towards developing neural network architectures \cite{simonyan2014very, he2016deep, hu2018squeeze, pmlr-v97-tan19a}. Image classification became the key task because of its appeal and challenges. While CE loss has been the most common loss function for this application from the past \cite{lecun1998gradient} to the present \cite{woo2023convnext}, other important applications such as face recognition have drawn researchers' attention to developing alternative loss functions \cite{koch2015siamese, wen2016discriminative, schroff2015facenet, liu2016large}. The CE loss function is applied at the end of the network where the predicted label is generated. Although this provides a good representation space in the penultimate layers, it is not sufficient for applications such as face recognition, which it is necessary for the features to be not only separable but also discriminative \cite{wen2016discriminative}. As a result, ideas such as Siamese networks \cite{koch2015siamese, Bromley1993SignatureVU}, Center Loss \cite{wen2016discriminative}, and Triplet Loss \cite{schroff2015facenet} emerged. 

The idea behind Siamese networks and Triplet Loss, which aim to create contrast between representations, is conceptualized under the term contrastive learning. However, contrastive learning has gained significant popularity \cite{chen2020simple, khosla2020supervised, he2020momentum, chen2020improved, radford2021learning} by addressing the problem of how to use unlabeled data in supervised tasks such as image classification (as opposed to tasks like super-resolution, which are inherently self-supervised).

FF \cite{hinton2022forward} and its developments \cite{lee2023symba, lorberbom2024layer, zhu2022contrastive, ahamed2023ffcl, scodellaro2023training, papachristodoulou2024convolutional, aghagolzadeh2024marginal} place the loss function after each layer. Although this is not equivalent to the contrastive ideas that apply the loss function in the penultimate layers, it similarly organizes the representation space; with the argument that the output of each layer can be considered as a representation space.

\paragraph{Contrastive learning.} The origin of the idea of contrastive learning can be attributed to \cite{Bromley1993SignatureVU}, where a Siamese network was used based on a distance metric between two representations generated from images of signatures, although at that time it was not referred to contrastive learning. Later, in \cite{hadsell2006dimensionality}, contrastive loss was used for dimensionality reduction. However, the major popularity of contrastive learning has come from pre-training networks in a self-supervised manner \cite{chen2020simple, he2020momentum, chen2020improved, radford2021learning}. In \cite{NIPS2014_07563a3f, xie2020unsupervised}, the idea was introduced that in an unsupervised task, although the labels of images are unknown, knowing that an image and its augmented version belong to the same class can allow the network to be trained in a self-supervised way. Subsequently, such developments, along with the introduction of a generalization of triplet loss \cite{schroff2015facenet} to utilize all non-matching samples \cite{sohn2016improved}, resulted in the highly popular method named SimCLR \cite{chen2020simple}. SimCLR is a simple, straightforward, and effective approach that trains the network in a self-supervised manner by applying contrastive loss in the representation space. This method proposes two forward passes through the network, where the inputs for each of these forward passes are different augmentations of the same batch. In this case, the corresponding images between these two forward passes are considered positive pairs, and the rest are considered negative pairs. In \cite{khosla2020supervised}, this method was even extended to the post-pretraining phase in a supervised manner, called SCL. In SCL, contrastive loss was adapted to a supervised version, where positive samples defined same class samples, and negative samples defined non-same class samples.
\paragraph{Forward-Forward.} In \cite{lee2023symba}, by symmetrizing the original FF loss function from the point of view of positive and negative samples, the convergence speed of training was increased. In \cite{lorberbom2024layer}, a mechanism was introduced to add collaboration between layers in FF. Although this improves the accuracy of the model, it eliminates the important property of FF, which is the independence of layers during the \textit{backward} stages. Along with FF, \cite{zhu2022contrastive} presented a method based on using Euclidean distance directly as local loss functions, grounded in contrastive learning in a self-supervised manner, which is used for pretraining the network. In \cite{ahamed2023ffcl}, the approach was used contrastive learning in a supervised manner by employing cosine similarity directly as the loss functions for CNNs. The use of this method within the framework of FF has been limited to pretraining before the network is fine-tuned using the BP on a specific application. From this viewpoint, this idea does not present a purely FF-based method and does not achieve satisfactory performance without the BP stage. Studies \cite{scodellaro2023training, papachristodoulou2024convolutional} also attempted to extend the FF to its application in CNNs. \cite{scodellaro2023training} focused on the effect of various CNN elements on FF, such as filter size, by evaluating only on the MNIST \cite{lecun1998gradient} dataset. In \cite{papachristodoulou2024convolutional}, by introducing a channel-wise competitive learning framework, the authors eliminated the need for negative data in the FF algorithm by defining a new loss function. In addition, this study proposed new convolutional blocks to better align CNNs with FF. To our best knowledge, no studies have been conducted to extend the use of FF to ViT. In our previous work \cite{aghagolzadeh2024marginal}, we introduced a new approach to FF named Contrastive Forward-Forward, focusing on Multi Layer Perceptron (MLP). In this study, with the same view of as our previous work, we extend it for use in ViT and provide a more comprehensive analysis.

\section{Method}
\label{sec:method}

This section begins by outlining the prerequisites for a better understanding of the proposed method, specifically focusing on FF \cite{hinton2022forward} and SCL \cite{khosla2020supervised} in their training and prediction phases. Next, it reviews the similarities and differences between these two algorithms. The following part presents the proposed method, which is a combination of both algorithms. Subsequently, it details how to apply FF and the proposed algorithm to ViT. Finally, we will introduce a modified loss function called Marginal Contrastive Loss.

\subsection{Preliminaries}

Consider \( f: \mathbb{R}^D \to \mathbb{R}^E \), where \( D \) is the dimension of the input sample, and \( E \) is the dimension of the representation formed by \( f \). Additionally, let \( x_i \) denote the \( i \)-th image of a given batch, and \( B \) be the number of samples within that batch. Let \( f(x_i) \) represent the embedding generated by a defined neural network. The neural network can be a single layer, multiple layers, or an encoder, depending on different configurations. For simplicity, we denote \( f(x_i) \) as \( f_i \). If there are two forward passes from the network in one step, we represent them as \( f_i^1 \) and \( f_i^2 \) respectively (or \( f_i^{pos} \) and \( f_i^{neg} \) depending on the content of the method).

\begin{figure}[t]
    \centering
    \begin{subfigure}[b]{0.16\columnwidth}
        \centering
        \includegraphics[width=\columnwidth]{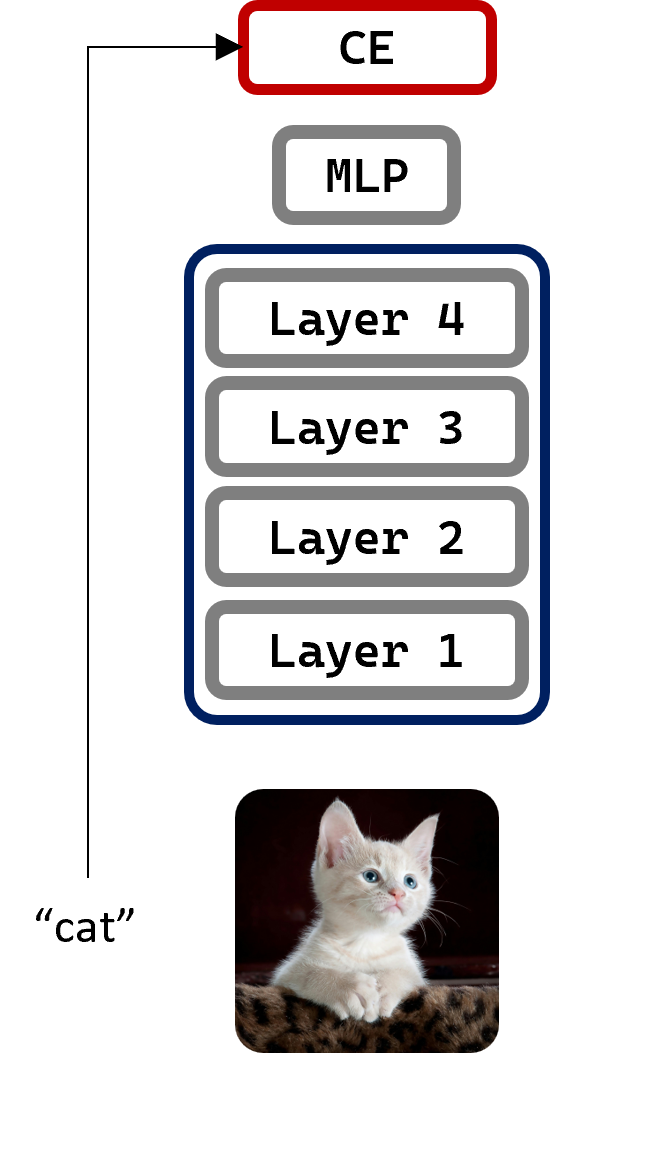}
        \caption{\scriptsize BP + CE}
        \label{fig:over1}
    \end{subfigure}
    \hfill
    \begin{subfigure}[b]{0.227\columnwidth}
        \centering
        \includegraphics[width=\columnwidth]{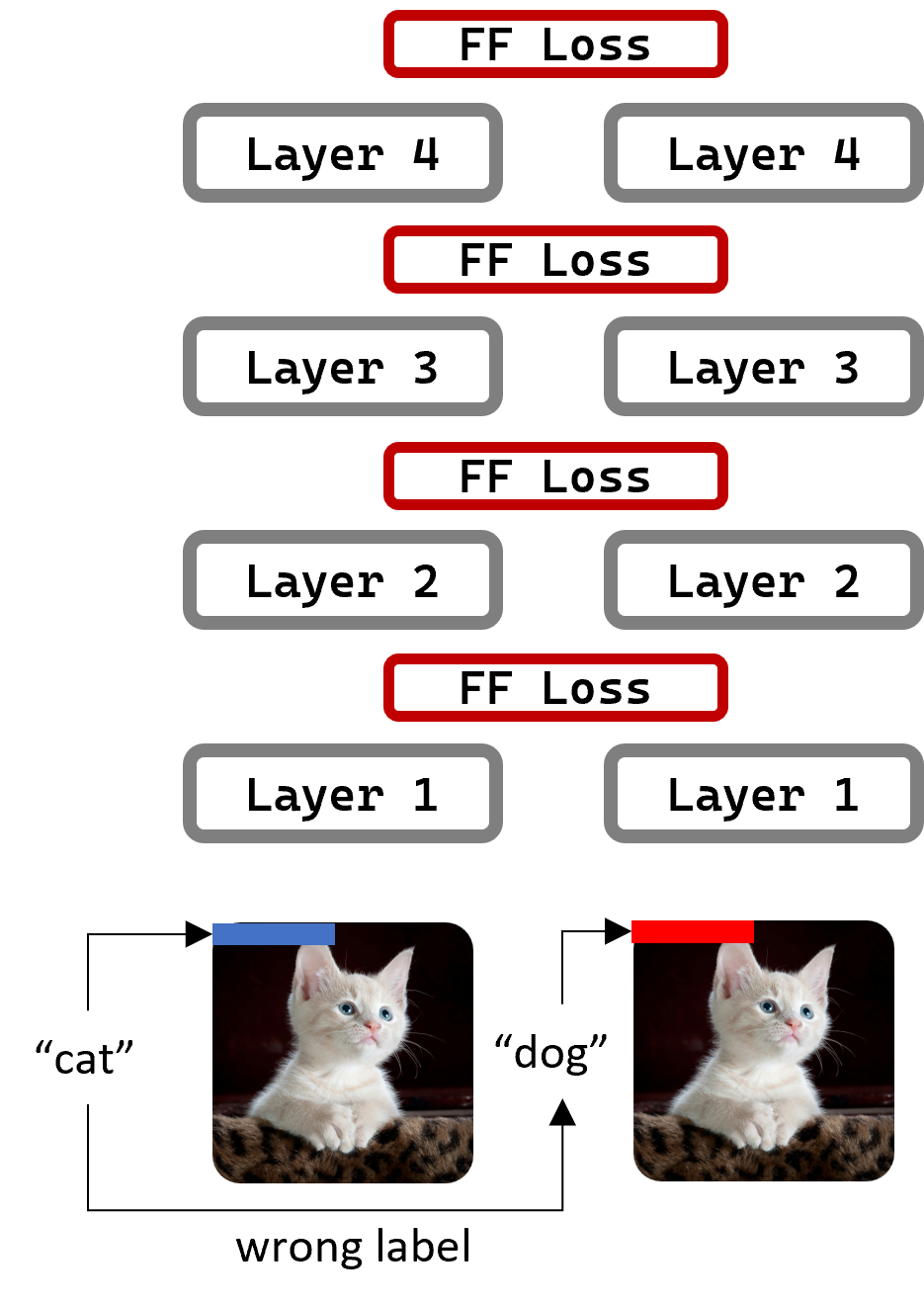}
        \caption{\scriptsize Baseline Forward-Forward}
        \label{fig:over2}
    \end{subfigure}
    \hfill
    \begin{subfigure}[b]{0.295\columnwidth}
        \centering
        \includegraphics[width=\columnwidth]{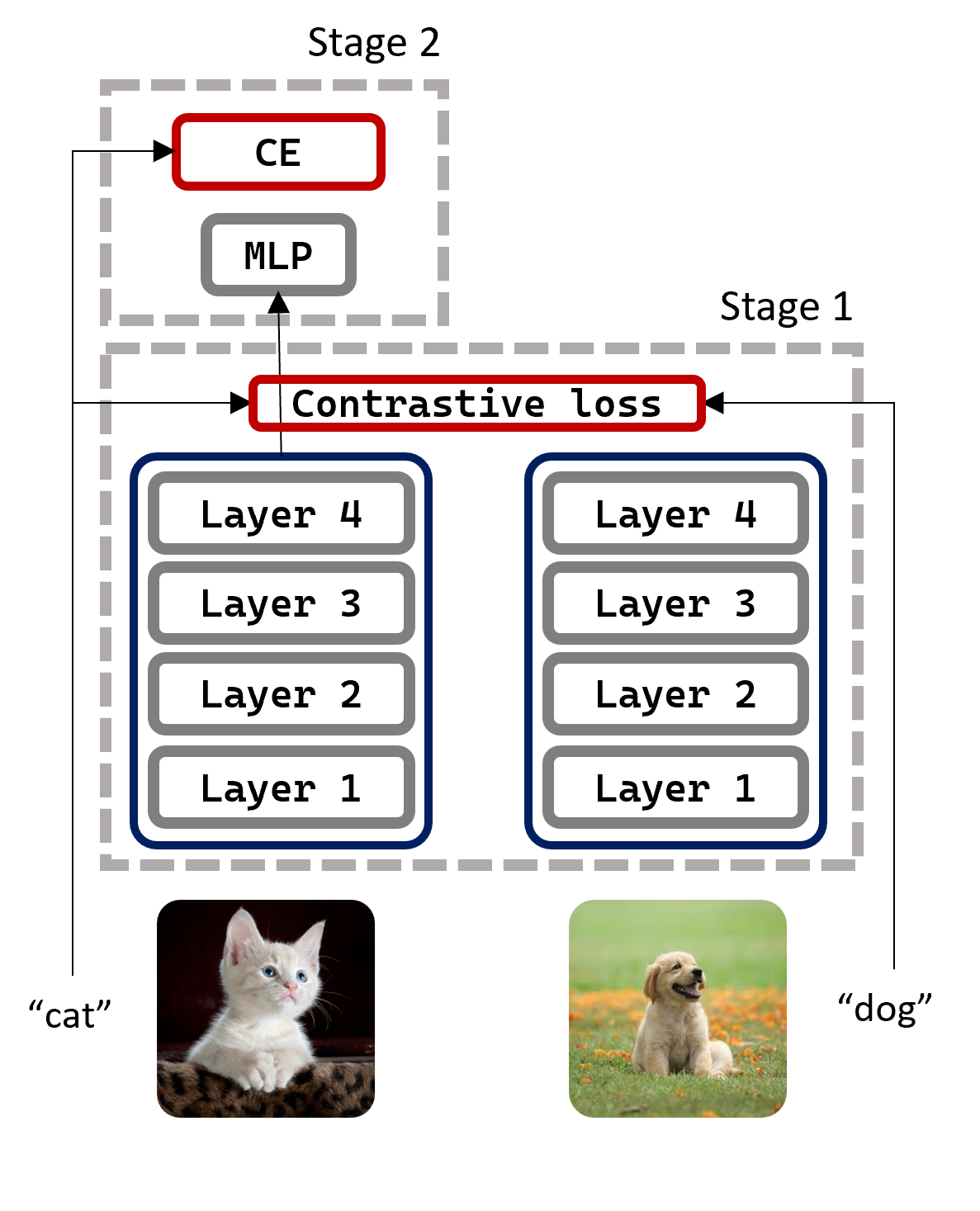}
        \caption{\scriptsize Supervised Contrastive Learning}
        \label{fig:over3}
    \end{subfigure}
    \hfill
    \begin{subfigure}[b]{0.27\columnwidth}
        \centering
        \includegraphics[width=\columnwidth]{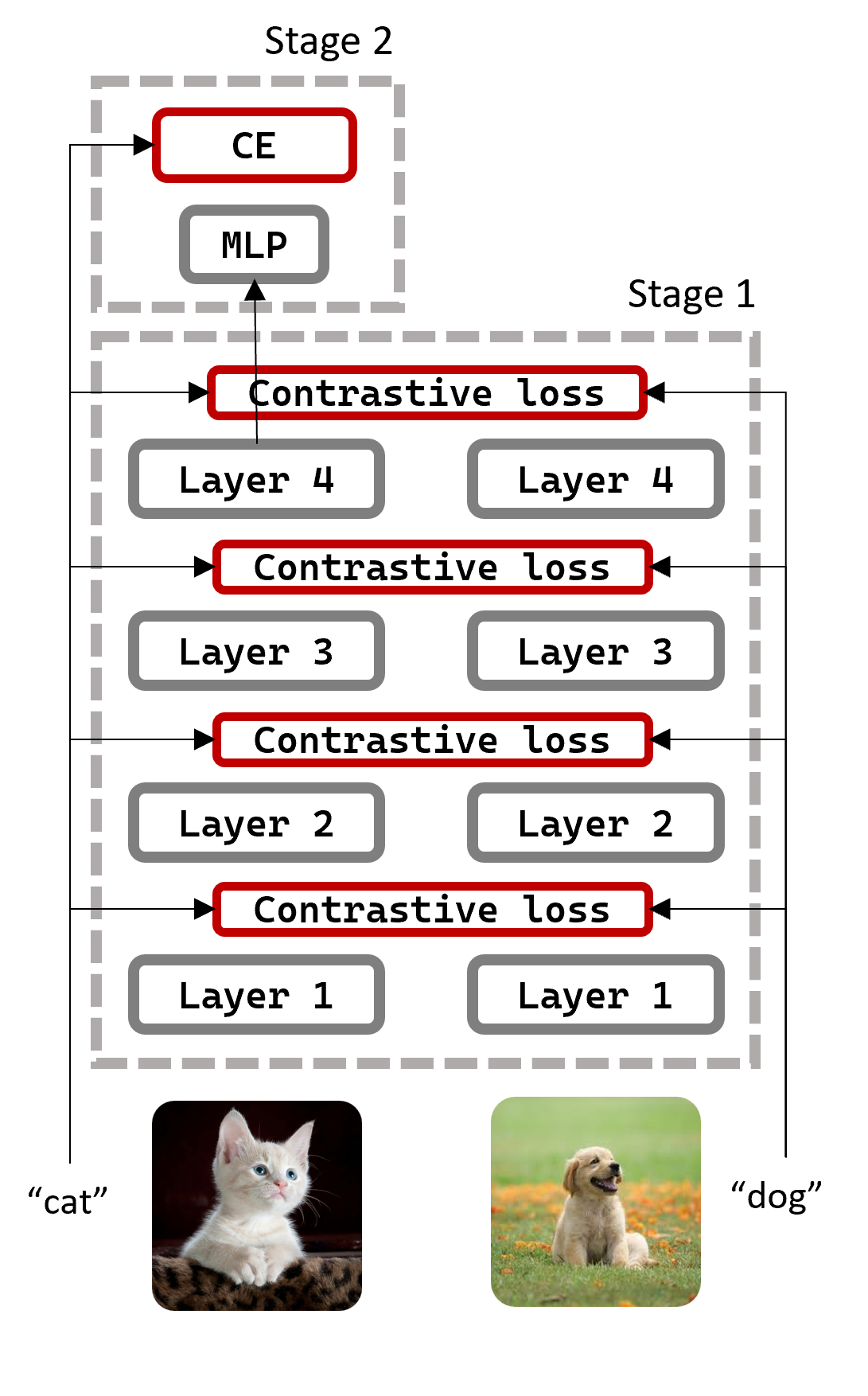}
        \caption{\scriptsize Contrastive Forward-Forward}
        \label{fig:over4}
    \end{subfigure}
    \caption{Abstract illustrations of backpropagation \cite{rumelhart1986learning} with Cross-Entropy loss function, baseline Forward-Forward \cite{hinton2022forward}, Supervised Contrastive Learning \cite{khosla2020supervised} and the proposed method, Contrastive Forward-Forward.}
    \label{fig:over}
\end{figure}

\subsubsection{Overview of Forward-Forward}
\paragraph{Training phase.}Unlike BP, which often uses a loss function at the end of the network, FF \cite{hinton2022forward} proposes to place the loss function after each layer, where the predicted label is not yet available. To this end, FF proposes a novel method for image classification. Initially, two new samples are generated for each image. The first sample is created by replacing the correct label, which is one-hot encoded, in the corner of the image, resulting in \( x_i^{pos} \). The second sample is obtained by replacing the incorrect one-hot encoded label in the corner of the image, referred to as \( x_i^{neg} \) (although, there is no insistence on this solution alone for integrating the label with the image). These two samples are then fed to the first layer through two forward passes to obtain \( {f_i^{pos}} \) and \( {f_i^{neg}} \). The local loss function in FF is defined as follows:

\begin{equation}
G_i^{pos} = \left\| {f_i^{pos}} \right\|_2^2 \quad G_i^{neg} = \left\| {f_i^{neg}} \right\|_2^2
\label{eq:goodness}
\end{equation}

\begin{equation}
L^{FF} = \sum\limits_{i = 1}^B {\log (1 + \exp (\theta  - G_i^{pos})}  + \log (1 + \exp (G_i^{neg} - \theta )
\label{eq:lossff}
\end{equation}

\( G_i \), representing the goodness of the \( i \)-th sample in batch, and \( \theta \), is a threshold parameter. The objective is to minimize this loss function, which occurs when \( G_i^{pos} \) increases and \( G_i^{neg} \) decreases. The addition of the threshold serves to further incentivize the model towards its goal. In the next step, \( f_i^{pos} \) and \( f_i^{neg} \), are used as inputs for the subsequent layers after normalization (i.e. for the next layer: \( x_i^{pos} = f_i^{pos} / \|f_i^{pos}\|_2 \) and \( x_i^{neg} = f_i^{neg} / \|f_i^{neg}\|_2 \)). Similarly, the process is repeated for the following layer. By defining such a loss function after each layer, the layers are encouraged to produce higher output values for images with correct labels and lower output values otherwise \cite{hinton2022forward}. Figs. \ref{fig:over1} and \ref{fig:over2} illustrate an abstract comparison between BP and FF.

\paragraph{Prediction phase.} Assuming that the layers have been trained to assign a higher goodness value when an image with the correct label is presented, it is sufficient to input an unseen sample with each possible label during prediction. The label that receives the highest score is selected. The score is calculated by summing the goodness values across all layers \cite{hinton2022forward}. One of the disadvantages of FF is that it requires C forward passes for each unseen sample in the prediction phase, where C is the number of classes. Although an alternative method was presented that performs inference with only a single forward pass \cite{hinton2022forward}, it leads to a drop in accuracy (as discussed in Sec. \ref{sec:exp3}).

\subsubsection{Overview of Supervised Contrastive Learning}
\paragraph{Training phase.}In the conventional network training method for image classification, CE loss function is placed at the end of the network to compare the predicted and true labels. SCL \cite{khosla2020supervised} proposes to place a loss function after the encoder to first organize the representation space of features (Fig. \ref{fig:over3}), such that samples of the same class are close together, while the samples from different classes are distanced from one another. To achieve this goal, the study suggests a batch of data undergoes two distinct paths of random augmentation to yield two augmented batches, \(x^1\) and \(x^2\). Then, these two batches pass through the encoder via two forward passes to produce \(f^1\) and \(f^2\) of size (\(B, E\)). If we concatenate them to form \(f = [f^1;f^2]\) (\(2B, E\) ), the loss function for these representations (or after passing through a projection layer \cite{khosla2020supervised}) is calculated as follows:

\begin{equation}
L^{\text{Supervised Contrastive}} = -\sum_{i=1}^{2B}  \frac{1}{|P(i)|} \sum_{p \in P(i)} \log \left( \frac{\exp(f_i \cdot f_p / \tau)}{\sum_{a \in A(i)} \exp(f_i \cdot f_a / \tau)} \right) 
\label{eq:supcon}
\end{equation}

Where $f_i$ called anchor. The set $P(i)$ includes the indices of representations within $f$ that belong to the same class as sample $i$, and \( |P(i)| \) is the cardinality of this set. \( A(i) \) contains the indices of all representations within \( f \) except \( i \). The dot product of \( a \) and \( b \) (\( a \cdot b \)) calculates the cosine similarity, assuming the vectors are normalized. Also, $\tau$ is a temperature parameter that controls the concentration level of the distribution \cite{oord2018representation,chen2020simple}. The loss function decreases when the similarity of the representation vectors of samples within the same class increases, and the similarity of the representation vectors of samples from different classes decreases.

Assuming the formation of the space, in the second stage (after training the encoder), the encoder weights are frozen, and a simple MLP network is used \cite{khosla2020supervised}. This MLP, usually just a linear layer, is trained using the CE loss function to map samples from the representation space created by the encoder to the correct labels (Fig. \ref{fig:over3} Stage 2).

\paragraph{Prediction phase.}Given that the last layer of the network contains an MLP mapping to labels, similar to conventional methods, in the prediction phase, a trained network in this manner maps an unseen sample to a label with a single forward pass.
\subsection{Conceptual Similarities: Forward-Forward and Supervised Contrastive Learning}
In FF, each data batch is fed into the network twice: once with the correct label and once with an incorrect label, across two forward passes. In SCL, two batches that have undergone two different and random augmentation paths are also fed into the network through two forward passes. In FF, the network weights are updated such that the input of an image batch with the correct label results in higher output values from the network layers, and lower output values for batches containing images with incorrect labels. Essentially, the network is trained to create a contrast between these two scenarios. SCL also pursues the creation of contrast with a different concept, i.e. aiming to establish this contrast by promoting similarity and dissimilarity among same and non-same class samples in the representation space, respectively. These similarities between the two methods inspired us to leverage insights from contrastive learning to modify FF.

\subsection{Proposed Method: Contrastive Forward-Forward}
\label{sec:cff}

We propose to replace the baseline loss function in FF with a contrastive loss function and adopt the data-feeding strategy of SCL by the idea of applying loss functions between layers (Fig. \ref{fig:over4}). We call this method Contrastive Forward-Forward (CFF).

\paragraph{Training phase.} 
A data batch undergoes two paths of random augmentation, resulting in two separate batches. These two batches are fed into the first layer, producing representations \(f^1\) and \(f^2\). Using Eq. \ref{eq:supcon} (or Eq. \ref{eq:msupcon}), the contrastive loss is calculated, and the weights of the first layer are updated accordingly. \(f^1\) and \(f^2\) are then input into the second layer. After the output is prepared  the  second loss is calculated, and the second layer's weights are updated. In the same way, it continues up for each layer until the end of the encoder. This process is repeated for all subsequent batches which can also be performed in parallel (Sec. \ref{sec:pip}) and then for the following epochs as well. After training the encoder, similar to the SCL strategy, with the frozen encoder, in the second stage a simple MLP network is employed. Training of this MLP is conducted using CE to learn how to map the representations to the correct labels.
\paragraph{Prediction phase.} Like SCL, despite the presence of an MLP layer that maps the representation to labels, for an unseen sample, the predicted label is obtained with a single forward pass.

\subsubsection*{CFF: A Generalization of FF}
\label{sec:wash}
In this section, we aim to demonstrate that the proposed method is a generalization of the baseline FF to use  images themselves as labels. To illustrate this, for simplicity, let each image be a vector with \( D \) elements, denoted by \( x_i \), where $i$ is the index of the sample in a batch. Also, consider that \( s_i \) denotes the label vector, which in the traditional method is a one-hot vector of length \( C \).

\begin{figure}[H]
    \centering
    \begin{subfigure}[b]{0.15\textwidth}
        \centering
        \includegraphics[width=\textwidth]{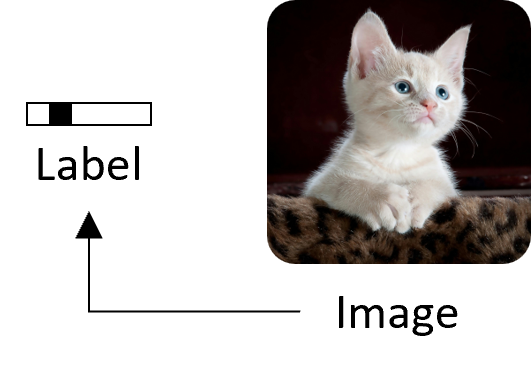}
        \caption{\scriptsize One-Hot vector, BP+CE}
        \label{fig:label1}
    \end{subfigure}
    \hfill
    \begin{subfigure}[b]{0.10\textwidth}
        \centering
        \includegraphics[width=\textwidth]{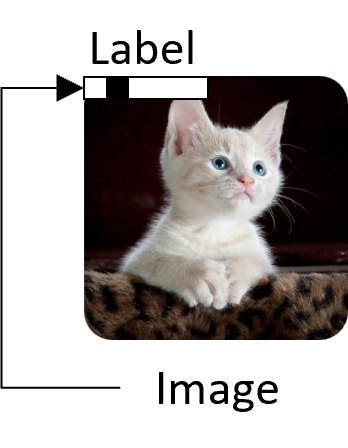}
        \caption{\scriptsize One-Hot on corner}
        \label{fig:label2}
    \end{subfigure}
    \hfill
    \begin{subfigure}[b]{0.197\textwidth}
        \centering
        \includegraphics[width=\textwidth]{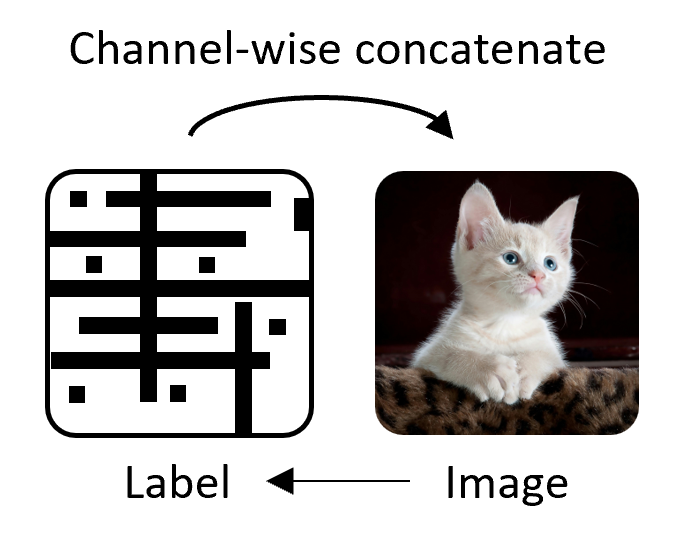}
        \caption{\scriptsize A random image size represention}
        \label{fig:label3}
    \end{subfigure}
    \hfill
    \begin{subfigure}[b]{0.17\textwidth}
        \centering
        \includegraphics[width=\textwidth]{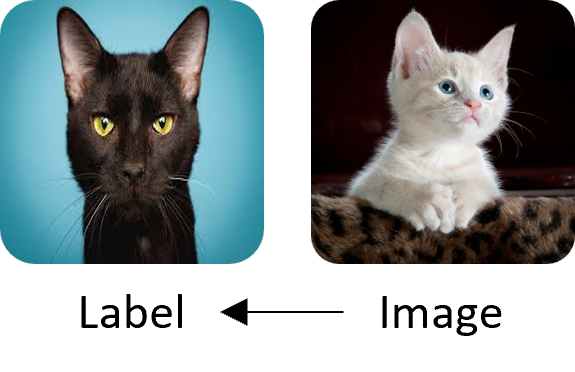}
        \caption{\scriptsize A certain image\\ ~~~~~~~~~~}
        \label{fig:label4}
    \end{subfigure}
    \hfill
    \begin{subfigure}[b]{0.262\textwidth}
        \centering
        \includegraphics[width=\textwidth]{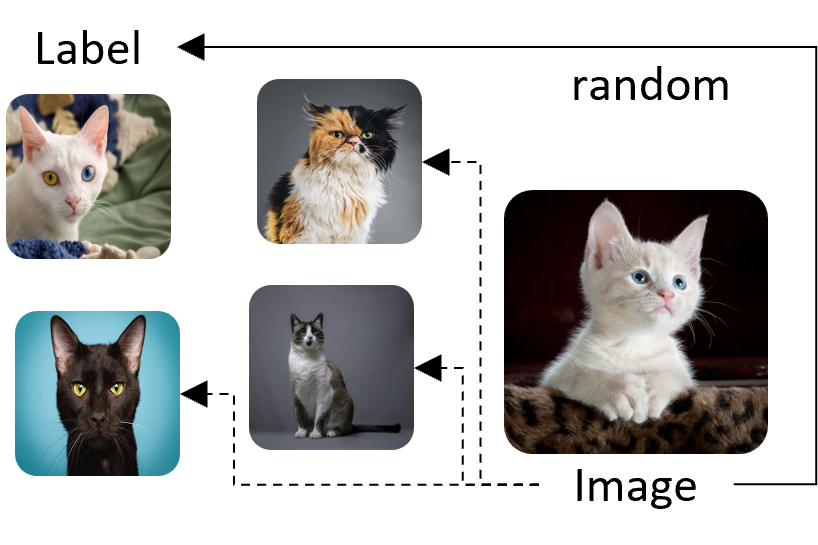}
        \caption{\scriptsize A random image\\ ~~~~~~~~~~}
        \label{fig:label5}
    \end{subfigure}
    \caption{Label Representation Strategies.}
    \label{fig:label}
\end{figure}

In BP with CE loss, the label is encoded with a one-hot vector and directly fed into the loss function (Fig. \ref{fig:label1}). In the original FF, the one-hot vector corresponding to the label places in a corner of the image (Fig. \ref{fig:label2}). Similar to \cite{lee2023symba} if we treat the label representation as a vector the same size as the image and concatenate it with the input, i.e. \(x_i^{\text{new}} =[x_i^{\text{old}}; s_i]\), the function \( f \) would be defined as: \(f: \mathbb{R}^{D+D} \to \mathbb{R}^{E}\). In this case, we have a label vector of length \( D \) for each class, while, FF had a one-hot vector. In the positive forward pass, the label representative for $i$'th sample is $s_i^{pos}$, and in the negative forward pass, the label representative for $i$'th sample is \( s_i^{neg}\), which is one of the \( C-1 \) other label vectors (randomly chosen). Thus, \( G_i^{\text{pos}} \) and \( G_i^{\text{neg}} \) will be written according to Eq. \ref{eq:goodness} as follows:

\begin{equation}
G_i^{pos} = \left\| {f([x_i;s_i^{pos}])} \right\|_2^2 \quad
G_i^{neg} = \left\| {f([x_i;s_i^{neg}])} \right\|_2^2
\end{equation}
Another idea that comes to mind is to pass \( x_i \) and \( s_i \) through \( f \) separately, i.e., obtain \( f(x_i) \) and \( f(s_i) \) (each \(f: \mathbb{R}^{D} \to \mathbb{R}^{E}\)). Then, we can define \( G_i \) based similarity as follows:

\begin{equation}
G_i^{pos} = f(x_i) \cdot f(s_i^{pos}) \quad
G_i^{neg} = f(x_i) \cdot f(s_i^{neg})
\label{eq:goodnessnew}
\end{equation}

For example, \( G_i^{\text{pos}} \) can be interpreted as the similarity between the representation produced by the specified sample and the representation of its corresponding class vector. \( C \) class vectors are available. In \cite{lee2023symba}, these class vectors were randomly initialized and fixed. However, instead of randomly choosing values for \( s \), is it not better to use the images themselves? That is, one image could be chosen as the representative for all images of the same class, and according to \ref{eq:goodnessnew}, the goodness increases when the representations are closer to this image based on dot similarity (or cosine similarity after normalization). In fact, instead of random initialization for \( s \) (Fig. \ref{fig:label3}), we are using one of the images (Fig. \ref{fig:label4}). Following this approach, we suggest going a step further and selecting a representative label randomly from the images corresponding to the class each time instead of choosing a fixed image (Fig. \ref{fig:label5}). In this case, Eq. \ref{eq:goodnessnew} can be rewritten as follows:

\begin{equation}
G_i^{pos} = f(x_i) \cdot f(x_{p\in P(i)}) \quad
G_i^{neg} = f(x_i) \cdot f(x_{n\in N(i)})
\label{eq:goodnessnew2}
\end{equation}
$P(i)$ contains the indices of the samples from the same class as $i$'th sample, excluding itself, and \( N(i) \) contains the indices of the samples from different classes.

Now consider the base FF loss presented in Eq. \ref{eq:lossff}. In \cite{lee2023symba}, the loss function was modified with the aim of symmetrizing it with respect to \( G_i^{pos} \) and \( G_i^{neg} \) as follows:

\begin{equation}
L = \sum\limits_{i = 1}^B {\log (1 + \exp (  \alpha(G_i^{neg} - G_i^{pos})))}
\label{eq:losssymba}
\end{equation}

Let's set \(\alpha\) to 1, this equation can be rewritten as follows:
\begin{equation}
L = \sum_{i=1}^{B} \log \left( 1 + \frac{\exp(G_i^{\text{neg}})}{\exp(G_i^{\text{pos}})} \right) =-\sum_{i=1}^{B} \log \left( \frac{\exp(G_i^{\text{pos}})}{\exp(G_i^{\text{pos}}) + \exp(G_i^{\text{neg}})} \right)
\end{equation}

If we substitute proposed $G_i$ using Eq. \ref{eq:goodnessnew2} and denote $f(x_i)$ in its simplified form as $f_i$ (and the other representations):

\begin{equation}
L=\sum_{i=1}^{B} -\log \left( \frac{\exp \left( f_{i} \cdot f_{p \in P(i)} \right)}{\exp \left( f_{i} \cdot f_{p \in P(i)} \right) + \exp \left( f_{i} \cdot f_{n \in N(i)} \right)} \right)
\end{equation}

According to the strategy, each label is randomly one of the images corresponding to its class each time. A better approach would be to involve all the samples in one step, such that:
\begin{equation}
E[\exp(f_i \cdot f_{p \in P(i)})] = \frac{1}{|P(i)|} \sum_{p \in P(i)} \exp(f_i \cdot f_{p})
\end{equation}
And similarly for the negative part, where \( E[\cdot] \) is the expectation operator. As a result:

\begin{equation}
L = \sum_{i=1}^{B} -\log \left( \frac{1}{|P(i)|} \sum_{p \in P(i)} \frac{\exp ( f_{i} \cdot f_{p})}{\frac{1}{|P(i)|} \sum_{p \in P(i)} \exp (f_{i} \cdot f_{p}) + \frac{1}{|N(i)|} \sum_{n \in N(i)} \exp (f_{i} \cdot f_{n})} \right)
\end{equation}

If we ignore the normalization coefficients in the denominator:

\begin{equation}
L = \sum_{i=1}^{B} -\log \left( \frac{1}{|P(i)|} \sum_{p \in P(i)} \frac{\exp ( f_{i} \cdot f_{p})}{ \sum_{a \in A(i)} \exp ( f_{i} \cdot f_{a})} \right)
\end{equation}

Where \( A(i) = P(i) \cup N(i) \). As a result, this equation exactly matches one of the forms of Supervised Contrastive Loss defined in \cite{khosla2020supervised} with \(\tau=1\). In another form, the summation over the positives is outside the logarithm which is shown in Eq. \ref{eq:supcon}. It has been shown experimentally that the second form performs better \cite{khosla2020supervised}. Therefore, we will also use the second form as the contrastive loss function.

\subsection{Contrastive Forward-Forward on Vision Transformers}\label{sec:cffvit}

\begin{figure}[H]
    \centering
    \includegraphics[width=1\linewidth]{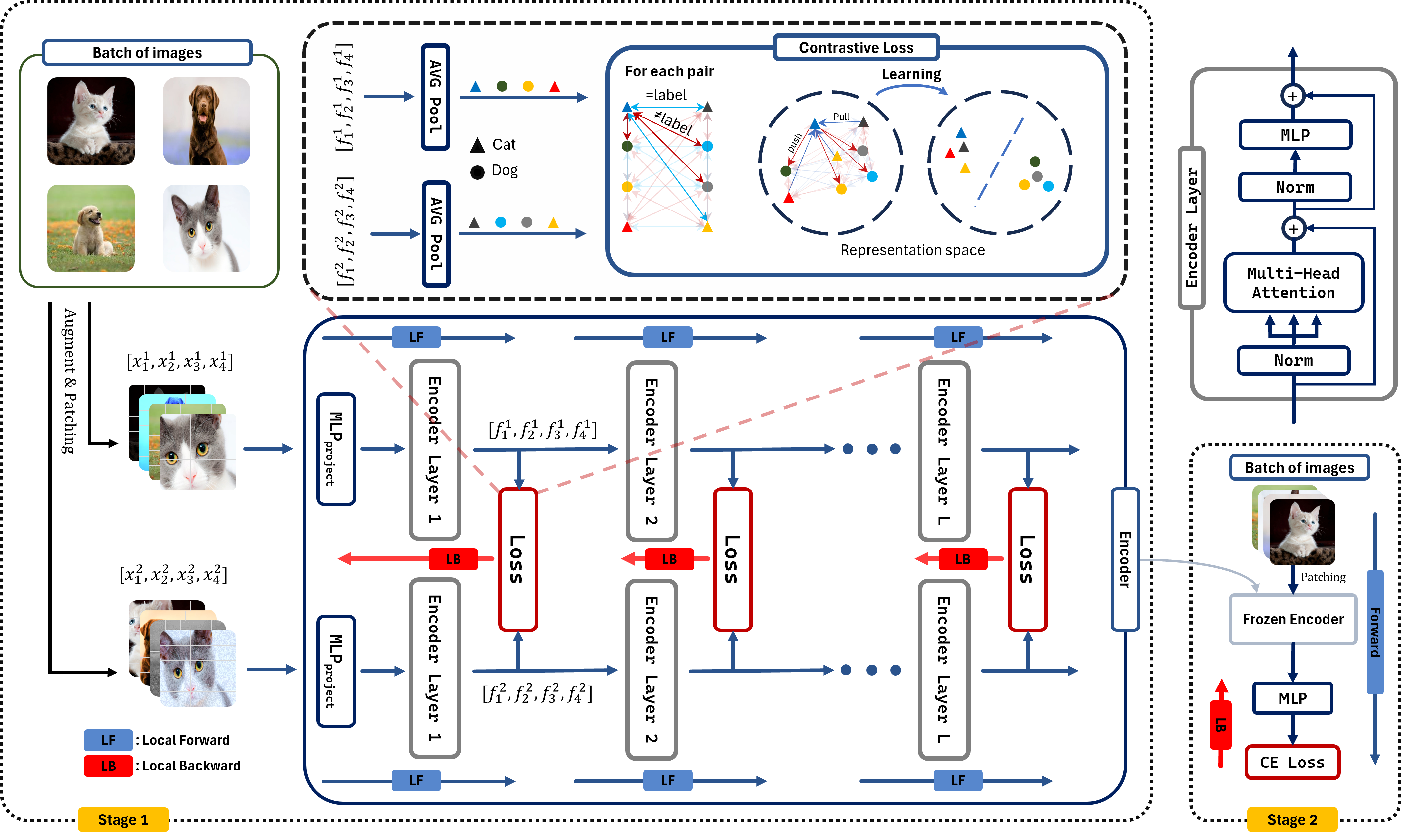}
    \caption{An illustration of the proposed training algorithm applied to ViT \cite{dosovitskiy2020image}. Each "Encoder Layer" is considered a layer of the network. For each layer, there are two local forward passes and one local backward pass. There is only one encoder in "Stage 1", and the bottom network is displayed for a better illustration of the algorithm; in fact, the layer weights in the corresponding layers of the top and bottom networks are shared.}
    \label{fig:overvit}
\end{figure}

Fig. \ref{fig:overvit} and Algorithm \ref{algo:proposed} (left side) represent our proposed method on ViT. The operation of projecting the patched image alongside one encoder layer is treated as the first layer from the proposed method perspective. After that, each encoder layer is considered as a separate layer. In particular, if we consider that \(x_i\) is a flattened patched sample with dimensions \((\# \text{patch}, h \times w \times c)\) where \(h\) and \(w\) are the height and width of the patch and \(c\) is the number of channels in the input image, the structure of the encoder (network without a classifier head) will be as follows:

\begin{flalign*}
f_{i,1} &= \text{encoder\_layer}_1 (\text{MLP}_{\text{proj}}(x_i) + E_{\text{pos}}) \quad & l = 1 \\
f_{i,l} &= \text{encoder\_layer}_l (f_{i,l-1}) \quad & l = 2, 3, \dots, L
\end{flalign*}
Such that each encoder\_layer\(_l\)(\(input\)) is defined as follows:
\begin{flalign*}
z_{i,l} &= \text{MSA}_l(\text{LN}(input_i)) + input_i \quad & l = 1, 2, \dots, L \\
f_{i,l} &= \text{MLP}_l(\text{LN}(z_{i,l})) + z_{i,l}
\end{flalign*}

Where L denotes the number of layers. This structure is consistent with the original ViT architecture \cite{dosovitskiy2020image}, including the use of Multi Self-Attention (MSA), Layer Normalization (LN) and positional encoding (\(E_{pos}\)) \footnote{For more details, please refer to \cite{dosovitskiy2020image}.}.

For image classification, in ViT, a patch called the class token, which has initial learnable values, is concatenated to the main image patches. After passing through the encoder, and assuming that it has been enriched, a MLP maps only this token to the predicted label \cite{dosovitskiy2020image}. An alternative approach does not use any additional tokens; instead, the features extracted from the encoder are passed through an average pooling layer and then fed to an MLP \cite{beyer2022better}. We chose the second approach for implementation in our proposed method and compared these two approaches in Sec. \ref{sec:ablative}. As shown in Fig. \ref{fig:overvit} similar to FF the loss function (Eq. \ref{eq:supcon} or \ref{eq:msupcon}) is applied after each layer, we pass every generated representation from that layer through an average pool to the loss function, but what is passed on to the next layer will be the same representation directly, except for the final layer, which is passed through an average pool before being delivered to a MLP. More precisely, consider the network notation as ViT[E H L], where E denotes the dimension of each embedded patch, H is the number of heads, and L is the number of transformer layers. Also, consider that \( f_{i,j,l} \) represents the representation generated from the \( j \)-th patch of the \( i \)-th sample at the output of the \( l \)-th encoder layer. In this case, the mentioned average pooling operation works as follows:

\begin{equation}
\hat{f}_{i,l} = \frac{1}{\#\text{patch}} \sum_{j=1}^{\#\text{patch}} f_{i,j,l}
\end{equation}

Compared to the proposed method, Algorithm \ref{algo:proposed} (right side) shows the baseline FF training approach for ViT with two modifications compared to the original FF \cite{hinton2022forward}. These modifications are: 1-concatenating the label as a patch (randomly initialized and fixed according to \cite{lee2023symba}) to the main image patches, instead of placing it as a one-hot vector and 2-feeding the mean of the representation along the tokens to the loss function, which does not change the fundamentals of original FF.

\begin{algorithm}[H]
    \definecolor{codeblue}{rgb}{0.25,0.5,0.5}
    \lstset{
      basicstyle=\fontsize{8.5pt}{8.5pt}\ttfamily\bfseries,
      commentstyle=\fontsize{8.5pt}{8.5pt}\color{codeblue},
      keywordstyle=\fontsize{8.5pt}{8.5pt},
    }
\begin{lstlisting}[language=python]
# This code includes the encoder's training stage related to each batch.
# x: features (B, D)
# y: targets  (B, 1)
# detach(): Blocks the gradient flow.
\end{lstlisting}
\begin{multicols}{2}
\begin{lstlisting}[language=python]
# Contrastive Forward-Forward
x1, x2 = augment(x), augment(x)
x1 = patching(x1) # (B, #P, D/#P)
x2 = patching(x2) # (B, #P, D/#P)
for l in range(L):
    f1 = layers[l].forward(x1) # (B, #P, E)
    f2 = layers[l].forward(x2) # (B, #P, E)
    f = cat(f1,f2) # (2B, #P, E)
    loss = ContrastiveLoss(f/norm2(f),y)
    loss.backward()
    optimizers[l].step()
    x1 = f1.detach()
    x2 = f2.detach()
\end{lstlisting}
\columnbreak 

\begin{lstlisting}[language=python]
# Forward-Forward
x, y_wrong = augment(x), wrong(y)
x = patching(x) # (B, #P, D/#P)
x1, x2 = cat_label_patch(x, y, y_wrong)
for l in range(L):
    f1 = layers[l].forward(x1) # (B, #P+1, E)
    f2 = layers[l].forward(x2) # (B, #P+1, E)
    loss = FFLoss(f1.mean(dim=1),
                  f2.mean(dim=1))
    loss.backward()
    optimizers[l].step()
    x1 = (f1/norm2(f1)).detach()
    x2 = (f2/norm2(f2)).detach()
\end{lstlisting}

\end{multicols}
\caption{\footnotesize PyTorch-style pseudocode for CFF (Proposed Method) and FF (baseline) on ViT}
\label{algo:proposed}
\end{algorithm}

\subsection{Marginal Contrastive Loss}
\label{sec:msupcon}
The Supervised Contrastive Loss \cite{khosla2020supervised}, in Eq. \ref{eq:supcon}, is minimized when there is maximized similarity between the representations of samples from the same class. In general, the capability of developing a representation with the whole encoder is higher than an individual layer of the encoder, so it is challenging to place this loss function at each layer. Since the first layers may not yet have access to high-level features, there is a potential risk of trapping in local minima. Furthermore, if the early layers create maximum proximity between samples of the same class with the features extracted up to that point, it leaves less room for adjustments in subsequent layers. To address this, we add a margin parameter to the contrastive loss function. This aims to gradually increase the closeness of representations of same-class samples from the first to the last layers. Given that \( A(i) = P(i) \cup N(i) \), where \( N(i) \) represents the set of indices that include non-same class samples to the \( i \)-th sample, Eq. \ref{eq:supcon} can be rewritten as Eq. \ref{eq:supconre}.

\begin{equation}
L^{\text{Supervised Contrastive}} = -\sum_{i=1}^{2B}  \frac{1}{|P(i)|} \sum_{p \in P(i)} \log \left( \frac{\exp(Q(f_i, f_p) / \tau)}{\sum_{p^\prime \in P(i)} \exp(Q(f_i, f_{p^\prime}) / \tau)+\sum_{n \in N(i)} \exp(f_i \cdot f_n / \tau)} \right) 
\label{eq:supconre}
\end{equation}
Such that \( Q(a, b) = a \cdot b \). In this case, we introduce the Marginal Contrastive Loss \footnote{"Supervised" has been omitted for simplicity in writing.} as shown in Eq. \ref{eq:msupcon}.

\begin{equation}
L^{\text{Marginal Contrastive}} =L^{\text{Contrastive}} \big|_{Q(a, b) = \min(a \cdot b + m, 1)}
\label{eq:msupcon}
\end{equation}

The representations are normalized before entering the loss function, resulting in their dot product yielding the cosine similarity. The value of the parameter $m$ ranges from 0 to 2. This parameter adds an artificial similarity, making two representations appear more similar than they are from the viewpoint of the loss function. The $\min$ operator is used to ensure that the similarity does not exceed 1. Adjusting the value of $m$ for each layer presents another challenge that will be addressed in experiments (Sec. \ref{sec:margintuning}).

To compute the gradient of this function, considering $\tau=1$, Eq. \ref{eq:msupcon} can be shown as follows:
\begin{equation}
L^{\text{Marginal Contrastive}}=\sum_{i=1}^{2B} L_i
\label{eq:losstotal}
\end{equation}
where:
\begin{equation}
L_i = \left(- \frac{1}{|P(i)|} \sum_{p \in P(i)} Q(f_i,f_p)\right) + \log \left( \sum_{p' \in P(i)} \exp(Q(f_i,f_{p'})) + \sum_{n \in N(i)} \exp(f_i\cdot f_n) \right)
\end{equation}

Consider \( f_k \) as the anchor, if \( r \in R(k) \) and \( r' \in R^c(k) \) such that \(P(k) = R(k) \cup  R^c(k)  \), and \( R(k) \) is a set containing the indices where \( f_k \cdot f_r > 1 - m \). In this case, the gradient of \( L_{k} \) with respect to \( f_k \) is equal to (further details are provided in Sec. \ref{sec:gradient}):

\begin{equation}
\frac{\partial L_k}{\partial f_k} = - \frac{1}{|P(k)|} \left( \sum_{r' \in R^c(k)} f_{r'} \right) + \frac{\sum_{r' \in R^c(k)} \exp(f_k\cdot f_{r'} + m) f_{r'} + \sum_{n \in N(k)} \exp(f_k\cdot f_n) f_n}{|R(k)| + \sum_{r' \in R^c(k)} \exp(f_k\cdot f_{r'} + m) + \sum_{n \in N(k)} \exp(f_k\cdot f_n)}
\label{eq:lossg}
\end{equation}

Gradient calculations in cases where \( f_k \) is not the anchor are provided in Eqs. \ref{eq:lossg2}, \ref{eq:lossg3}, and \ref{eq:lossg4}. Considering the update of a specific weight \( w \) in the local backward pass passes through the gradient of all present representation (\( \frac{\partial L_i}{\partial w} = \sum_{k=1}^{2B} \frac{\partial L_i}{\partial f_k} \times \frac{\partial f_k}{\partial w} \)). Given that Eqs. \ref{eq:lossg}, \ref{eq:lossg2}, \ref{eq:lossg3}, and \ref{eq:lossg4} (the local backward pass follows one of these equations depending on the scenario), for an arbitrary \( k \), the changes in the representations generated from \( r \in R(k) \) in the range \( 1-m < f_k \cdot f_r < 1 \) play no role in updating the weights through the representations of each other. While the entry and exit of representations into this range play a role in altering \( |R(k)| \) and \( |R^c(k)| \). This causes the method to focus more on correcting hard positive samples i.e. \( f_k \cdot f_{r^\prime} < 1 - m \) (\( r^\prime\in R^c(k)\)) while allowing the model flexibility to place \(  1-m < f_k \cdot f_{r} < 1 \) (\( r \in R(k) \)).

\section{Experiments}\label{sec:exp}
\begin{figure}[t]
    \centering
    \begin{subfigure}[b]{0.47\textwidth}
        \centering
        \includegraphics[width=\textwidth]{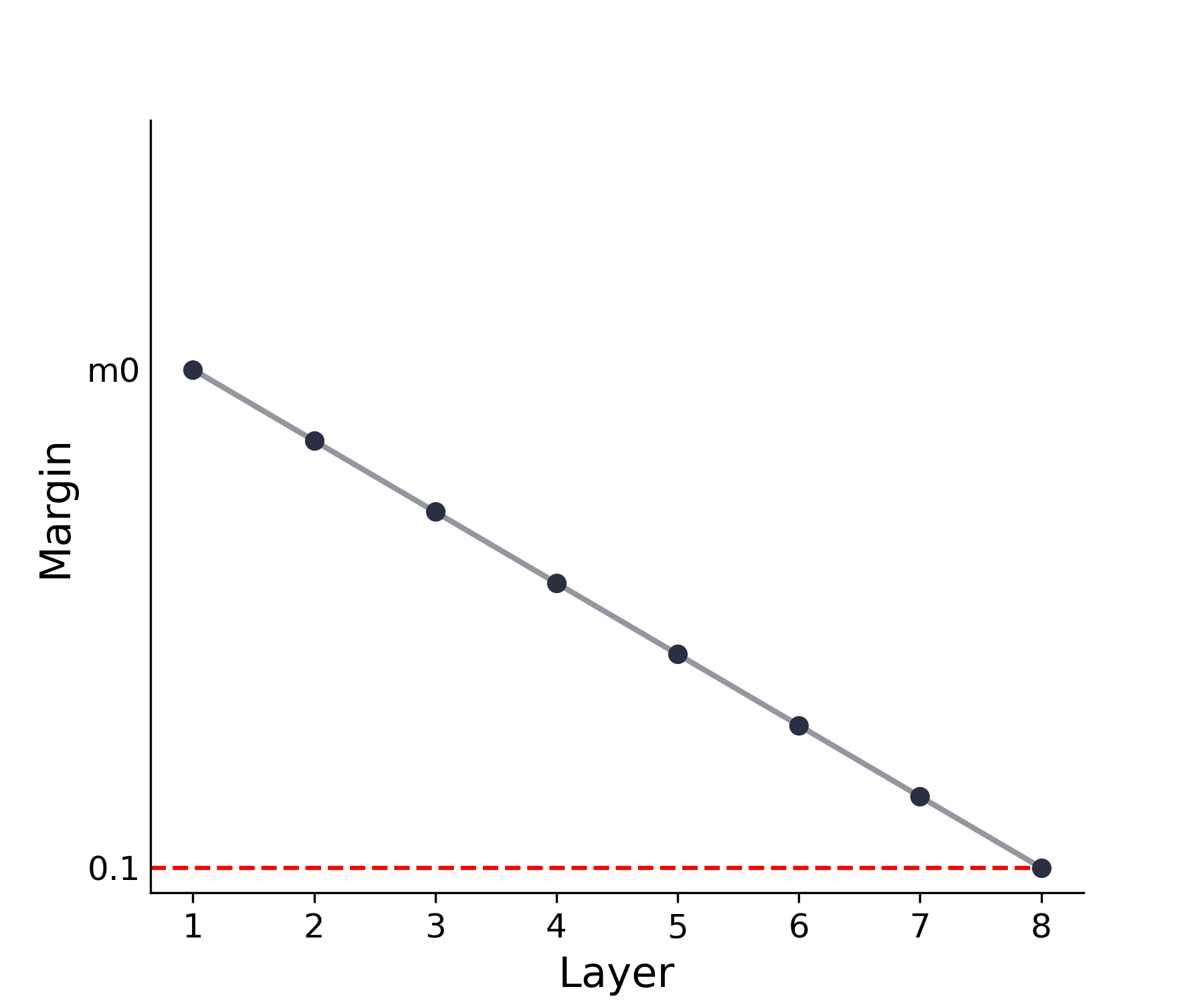}
        \caption{The strategy for setting margin values for each layer with \(L=8\)}
        \label{fig:sensitivitya}
    \end{subfigure}
    \hfill
    \begin{subfigure}[b]{0.47\textwidth}
        \centering
        \includegraphics[width=\textwidth]{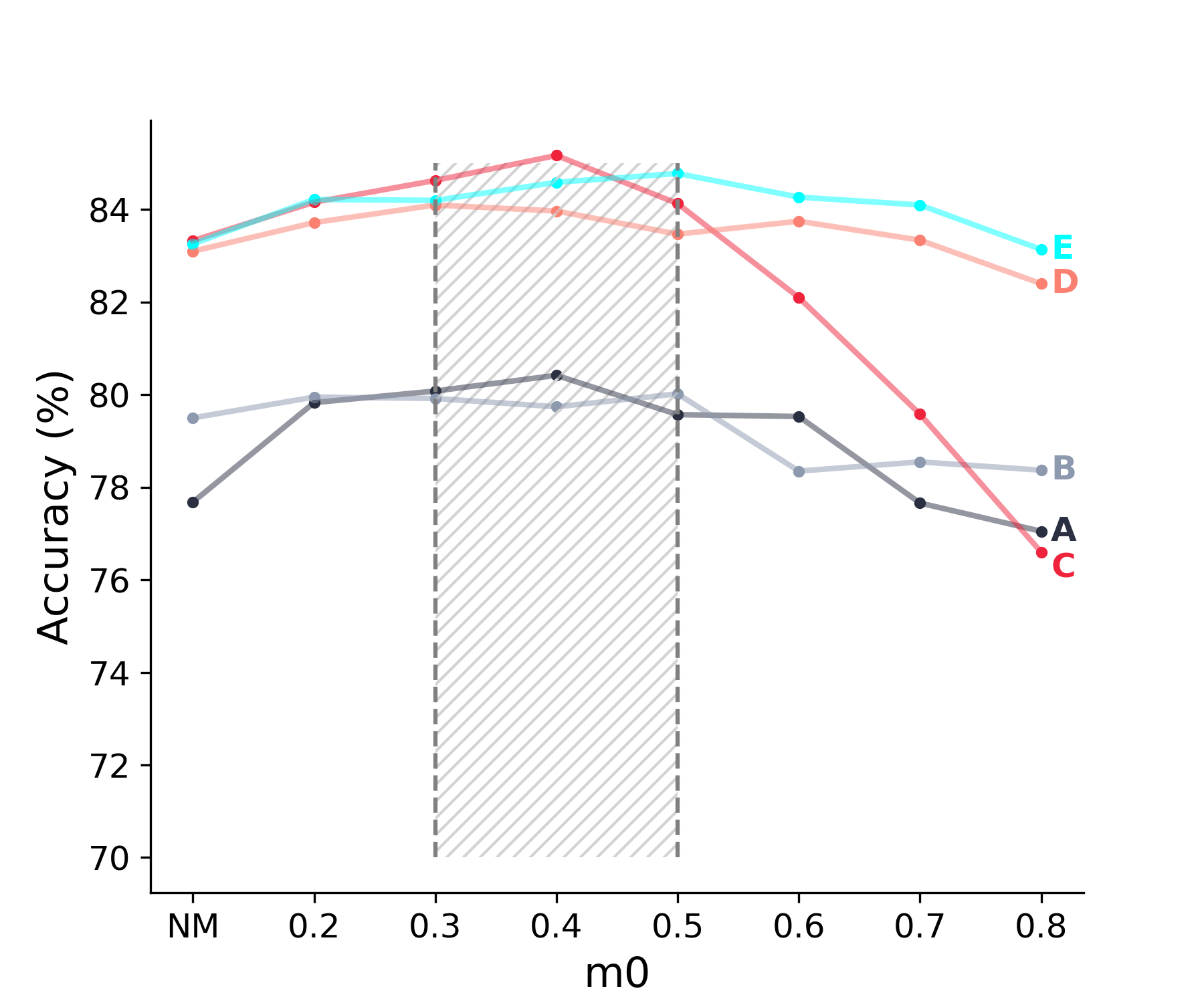}
        \caption{
Accuracy based on different $m_0$ values in various scenarios}
        \label{fig:sensitivityb}
    \end{subfigure}
    \caption{Margins tuning of proposed loss function with validation set, \textbf{A}: \textit{(ViT[128 4 5], CIFAR10, RC)}, \textbf{B}: \textit{(ViT[192 6 6], CIFAR10, RC)}, \textbf{C}: \textit{(ViT[192 6 6], CIFAR10, RC + RA)}, \textbf{D}: \textit{(ViT[192 6 6], CIFAR100, RC + RA)} and \textbf{E}: \textit{(ViT[240 6 7], CIFAR100, RC + RA)}, \textbf{NM}: No Margin, RC: Random Corp, RA: RandAug \cite{cubuk2020randaugment}}
    \label{fig:sensitivity}
\end{figure}

In this section, experiments are prepared to evaluate the proposed method. All experiments focus on the image classification problem. First, the process of tuning the margin hyperparameter in the Marginal Contrastive Loss was described (Fig. \ref{fig:sensitivity}), followed by a layer-wise analysis (Fig. \ref{fig:LW}). Then, the proposed method is compared with the baseline FF and some FF-based methods in terms of accuracy (Table \ref{tab:cffvsff}), inference time (Table \ref{tab:cffvsff_inference}), and convergence speed (Fig. \ref{fig:convergence}). The comparison with BP across various settings is then conducted (Tables \ref{tab:cffvsbp} \& \ref{tab:noise} and Fig. \ref{fig:complexity}). Finally, the contribution of the components to the proposed method is analyzed (Tables \ref{tab:tokenvspool} \& \ref{tab:onevstwo}). The experiments were conducted using 4 public benchmark datasets: MNIST \cite{lecun1998gradient}, CIFAR-10 \cite{krizhevsky2009learning}, CIFAR-100 \cite{krizhevsky2009learning} and Tiny ImageNet \cite{le2015tiny}. A description of these four datasets is provided in Sec. \ref{sec:Adataset}. In all scenarios, 10\% of the training data was randomly selected as validation data. During training, the weights for a specific model were chosen based on the best performance with respect to the loss obtained from the validation set. Consider that the test set of the datasets used has already been separated by the provider. In all experiments focusing on accuracy comparison, TOP-1 accuracy is used for MNIST and CIFAR-10, TOP-5 accuracy for CIFAR-100, and TOP-10 accuracy for Tiny ImageNet (all have been reported on the test set) are used. The proposed method, i.e., Contrastive Forward-Forward, is denoted as CFF (with baseline Contrastive Loss, Eq. \ref{eq:supcon}) and CFF+M (with Marginal Contrastive Loss, Eq. \ref{eq:msupcon}).

\paragraph{Models.} In a subset of experiments, a network consisting solely of an MLP was used, with its structure specified as MLP[E L], where L and E are the number of layers and the number of neurons (embedded feature) in each layer, respectively. For example, a 4-layer MLP with 800 units per layer is denoted as MLP[800 4]. ReLU \cite{Nair2010RectifiedLU} is used as the activation function after layers. The ViT models follow the same structure described in Sec. \ref{sec:cffvit} denoted as ViT[E H L]. For simplicity, as in \cite{beyer2022better}, learnable values were used for positional encoding \(E_{pos}\). In methods that require a classifier head, a linear projection head with the number of units equal to the number of classes is placed at the end of the network. AdamW \cite{DBLP:conf/iclr/LoshchilovH19} was used as the optimizer for all experiments \footnote{Further details of training settings are provided at: \href{https://github.com/HosseinAghagol/ContrastiveFF}{github.com/HosseinAghagol/ContrastiveFF}}.

\subsection{Margins Tuning}
\label{sec:margintuning}
The main challenge in using the proposed modified contrastive loss i.e. Marginal Contrastive Loss is tuning its hyperparameter, the margin. This hyperparameter can be set to different values for each layer in the proposed method. The goal of the proposed method is to increase the similarity between samples of the same class and decrease the similarity between samples of different classes in the final layer. The main idea behind using the margin is to achieve this gradually across the layers. Therefore, the margin value is considered to reduce from a higher value in the first layers to a lower value in the last layer. For simplicity, we considered this process linearly. Specifically, the margin value for the first layer is $m0$, and it linearly decreases to 0.1 in the final layer. This scenario is illustrated in Fig. \ref{fig:sensitivitya}. Therefore, for the entire network, it is only necessary to find an appropriate value for $m0$. Although we believe that adopting more complex approaches, such as non-linear reduction or tuning each margin individually, could lead to better performance, simplicity is prioritized. In Fig. \ref{fig:sensitivityb}, $m_0$ is considered in various values ranging from 0.2 to 0.8. According to the figure, in different scenarios, including different networks and types of augmentations on the CIFAR-10 and CIFAR-100 datasets, setting the value of $m_0$ between 0.3 and 0.5 generally yields good results. The midpoint of this range, 0.4, is found to be desirable and is chosen for all subsequent experiments.

\subsection{Layer-wise Training Analysis}
\begin{figure}[t]
  \centering
  \includegraphics[width=1\linewidth]{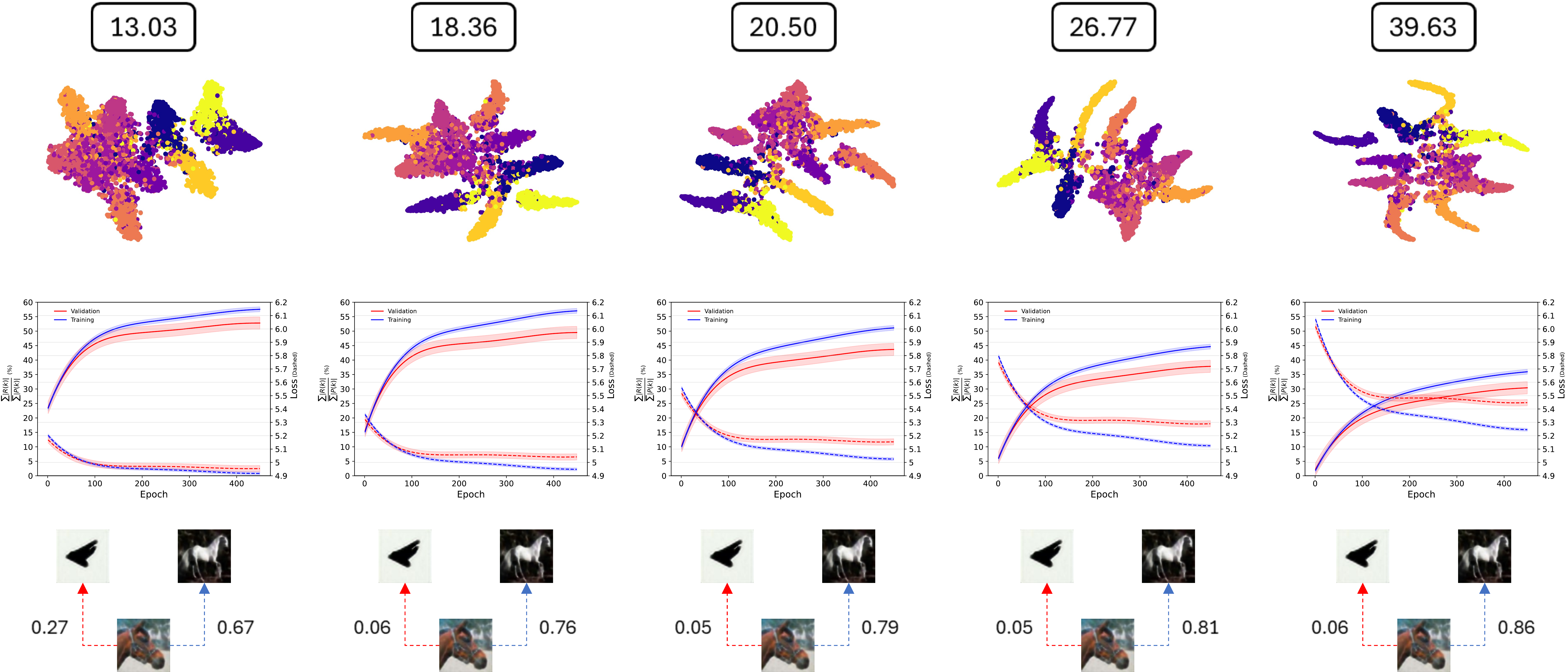}
  \caption{A layer-wise analysis of the training process of the proposed method (CFF+M) with ViT[128 4 5] on CIFAR-10. The first to fifth columns represent the first to fifth layers of the network, respectively. Row 1: The value of the Fisher criterion after training on the representations outputted from each layer with validation data. Row 2: 2D visualizations of the representations from each layer with validation data using t-SNE transformation. Row 3: The right vertical axis shows the loss, and the left vertical axis shows the percentage of samples from \(P(k)\) that fall into \(R(k)\). The summation operation is performed over each possible \( k \), which includes indices of all samples in the batch. The values of both charts are derived from the average of ten trials. Row 4: The cosine similarity between an arbitrary anchor and two positive and negative samples.}
  \label{fig:LW}
\end{figure}

Placing the loss function after each layer can improve the interpretability of the training process. Fig. \ref{fig:LW} attempts to illustrate what happens during the learning process of the proposed method with the marginal contrastive loss. The first row of this figure shows the value of the Fisher's linear discriminant criterion \cite{fisher1936use} after training for each layer. More precisely, this criterion was calculated on the features passed through average pooling after each layer for the validation data, as follows:
\begin{equation}
S_w = \sum_{c=1}^{C} S_c \quad , \quad S_B = \frac{1}{N} \sum_{c=1}^{C} N_c (m_c - m)(m_c - m)^T
\end{equation}
Where
\begin{equation}
m = \frac{1}{N} \sum_{i=1}^{N} f_i \quad , \quad m_c = \frac{1}{N_c} \sum_{i \in (label=c)} f_i
\end{equation}
\begin{equation}
S_c = \frac{1}{N} \sum_{i \in (label=c)} (f_i - m_c)(f_i - m_c)^T
\end{equation}
\(N\) is the total number of samples, \(N_c\) is the number of samples in class \(c\), \(m\) is the overall mean of the representations, \(m_c\) is the mean of the representations within class \(c\), \(S_c\) is the within-class scatter matrix for class \(c\), representing the spread of representations within each class, \(S_W\) is the total within-class scatter matrix, representing the combined scatter within all classes and \(S_B\) is the between-class scatter matrix, representing the variance between the means of the different classes. In this case, the Fisher criterion in the multi-class scenario is defined as follows.

\begin{equation}
Fisher~criterion = \text{Tr} \left\{ S_w^{-1} S_B \right\}
\end{equation}

The operator \( \text{Tr} \{.\} \) calculates the trace of a matrix.
In Fig. \ref{fig:LW}, the values of this criterion for the representations generated after each layer increased from around 13 to 40.
The second row of Fig. \ref{fig:LW} shows the 2D visualizations resulting from the t-SNE transformation \cite{JMLR:v9:vandermaaten08a} on the representations outputted from each layer, providing a qualitative assessment that confirms the results observed with the Fisher criterion. In the next row of this figure, there are graphs that have two vertical axes. The right vertical axis in each graph corresponds to the training and validation loss values for each epoch. Although the last layers perform better in terms of the Fisher criterion and same-class sample proximity in the qualitative charts due to receiving richer features, they are faced with higher loss due to the lower margin in the loss function. In fact, this can be considered as an advantage, as the last layers are encouraged more than expected.

As mentioned in Sec. \ref{sec:msupcon}, recall that \( R(k) \subseteq P(k) \) as the set of same-class sample indices for the \( k \)-th sample that satisfy the condition \( f_k\cdot f_r > 1 - m \) where \(r \in R(k)\), then \( \min(f_k\cdot f_r + m, 1) = 1 \). The left vertical axis in the third-row graphs shows the sum of \( R(k) \) for all possible \( k \)'s. This count is normalized by the number of samples in \( P(k) \) to represent the percentage of samples that fall within the set \( R(k) \) during training. These graphs also show that, although a higher percentage of cases should fall within this set in the last layers due to richer features, the lower margin in the final layers encourages these layers to work harder to bring same-class samples closer together. Interestingly, the behavior of these graphs is very similar to the loss graphs, as the performance gap between validation and training data is discernible in both sets of graphs. In fact, this criterion, like the loss graph, can reveal the occurrence of overfitting on last epochs. This means that the proposed method can add an additional criterion for interpretability of the network's training process.

In the last row of Fig. \ref{fig:LW}, the cosine similarity between an anchor and two samples (one from the same class and one from a different class) is shown, illustrating how this similarity gradually moves in the correct direction through the layers.

\subsection{Improving Performance with Contrastive Forward-Forward}
\label{sec:exp3}

The experimental results of the proposed method are compared with baseline FF \cite{hinton2022forward} and some FF-based methods i.e. SymBa \cite{lee2023symba} and FF Collab \cite{lorberbom2024layer} across different models and datasets are shown in Tables \ref{tab:cffvsff} and \ref{tab:cffvsff_inference}. In all experiments in this section, Random Crop was used as the augmentation method (except for MNIST). Our proposed algorithm outperforms the other algorithms in all scenarios except on MNIST which is close. Both SymBa and FF collab, which were introduced to enhance FF in MLPs, showed better performance compared to FF on this model. However, their performance on ViT was relatively weaker. Furthermore, our proposed loss function for use in the proposed algorithm, named Marginal Contrastive Loss (abbreviated as CFF+M), outperforms  the original contrastive loss \cite{khosla2020supervised} in the proposed method.

Our proposed method also performed better in terms of inference time (Table \ref{tab:cffvsff_inference}), as expected, due to the need for only one forward for an unseen sample, while FF requires C forward passes. However, there is a solution to match the inference time by adding a second stage to the model trained with FF, SymBa and FF Collab. In this second stage, a linear projection layer is added at the end of the network, with input from the outputs of all layers except the first layer \cite{hinton2022forward}. Then, for each sample, a label representative (a full zeros one-hot size vector in original FF or a full zero patch in our extension of FF to ViT) is input to the network along with the image. Then, only this head is trained with CE loss. Specifically, with this layer, for an unseen sample, a single forward pass is sufficient to predict the label. The result of this approach (adding the head) is also presented in Table \ref{tab:cffvsff_inference}, which, although achieving a desirable inference time, has significantly reduced accuracy.

\begin{table}[H]
\footnotesize
    \centering
    \caption{Performance comparison of Contrastive Forward-Forward (CFF \& CFF+M) against Baseline Forward-Forward (FF) and other FF-based methods on various datasets with MLP and ViT networks in terms of accuracy.\\}
    \begin{tabular}{l l c c c c c}
    \toprule
    Dataset & Model & FF\textsuperscript{\(\star\)}  & SymBa\textsuperscript{\(\dag\)}  & FF Collab\textsuperscript{\(\dag\)}  & CFF (\scriptsize{\textit{ours}}) & CFF+M (\scriptsize{\textit{ours}}) \\
    \midrule
    MNIST & MLP[500 3] 
        & 97.12
        & \textbf{97.82} 
        & \textbf{97.70} 
        & 96.33 
        & \textbf{97.69} \\
    CIFAR10 & MLP[800 4] 
        & 50.35
        & 56.80 
        & 54.33 
        & 54.32 
        & \textbf{57.33} \\
    CIFAR10 & ViT[128 4 5] 
        & 76.21 
        & 69.13 
        & 70.66
        & 78.70 
        & \textbf{80.42} \\
    CIFAR100 & ViT[192 6 6] 
        & 72.15
        & 68.78 
        & 71.79
        & 79.23
        & \textbf{80.39} \\
    TImagenet & ViT[240 6 7] 
        & 62.11
        & 59.46 
        & 62.15 
        & 72.43 
        & \textbf{73.23} \\
    \bottomrule
    \end{tabular}
    \label{tab:cffvsff}
\end{table}

{\tiny\(\star\)The official FF code has been written on the Matlab platform and applied to a MLP and a sparse MLP \cite{hinton2022forward}. The results obtained from this method are the product of our implementation on Python, which aligns with unofficial implementations. Additionally, the extension of this method for use in ViT has been carried out by us.

\(\dag\)There is no official code available for these methods, and their implementation has been done based on the description provided in the reference papers \cite{lee2023symba,lorberbom2024layer}, and the extension of these methods for use in ViT, similar to FF, has been carried out according to our proposed approach.
}

\begin{table}[H]
\footnotesize
    \centering
    \caption{Performance comparison of FF and other FF based methods with or without head against proposed method on CIFAR-10 with ViT in terms of accuracy (Acc.) and inference time (Infer. T) for an unseen sample. Having a head refers to a fully connected layer with SoftMax placed at the end of the network, enabling the network to predict an unseen sample in a single forward pass (corresponding to "one-pass softmax" testing procedure in \cite{hinton2022forward}).\\}
    \begin{tabular}{l c c c}
    \toprule
    Algorithm & Head & Acc.(\%) & Infer. T (ms$\times10$) \\
    \midrule
    \multicolumn{4}{c}{CIFAR10, ViT[128 4 5]}\\
    \midrule
    \multirow{2}{*}{FF} & - & 76.21 & 10.0 \\
    & \checkmark & 65.33 & 1.2 \\
    \midrule
    \multirow{2}{*}{SymBa } & - & 69.13 & 10.0 \\
    & \checkmark & 60.15 & 1.2 \\
    \midrule
    CFF+M (\scriptsize{\textit{ours}}) & \checkmark & \textbf{80.19} & \textbf{1.0} \\
    \midrule
    \multicolumn{4}{c}{CIFAR100, ViT[192 6 6]}\\
        \midrule
    \multirow{2}{*}{FF} & - & 72.15 & 110.0 \\
    & \checkmark & 62.77 & 1.3 \\
    \midrule
    CFF+M (\scriptsize{\textit{ours}}) & \checkmark & \textbf{80.39} & \textbf{1.1} \\
    \bottomrule
    \end{tabular}
    \label{tab:cffvsff_inference}
\end{table}

\begin{figure}[H]
  \centering
   \includegraphics[scale=0.32]{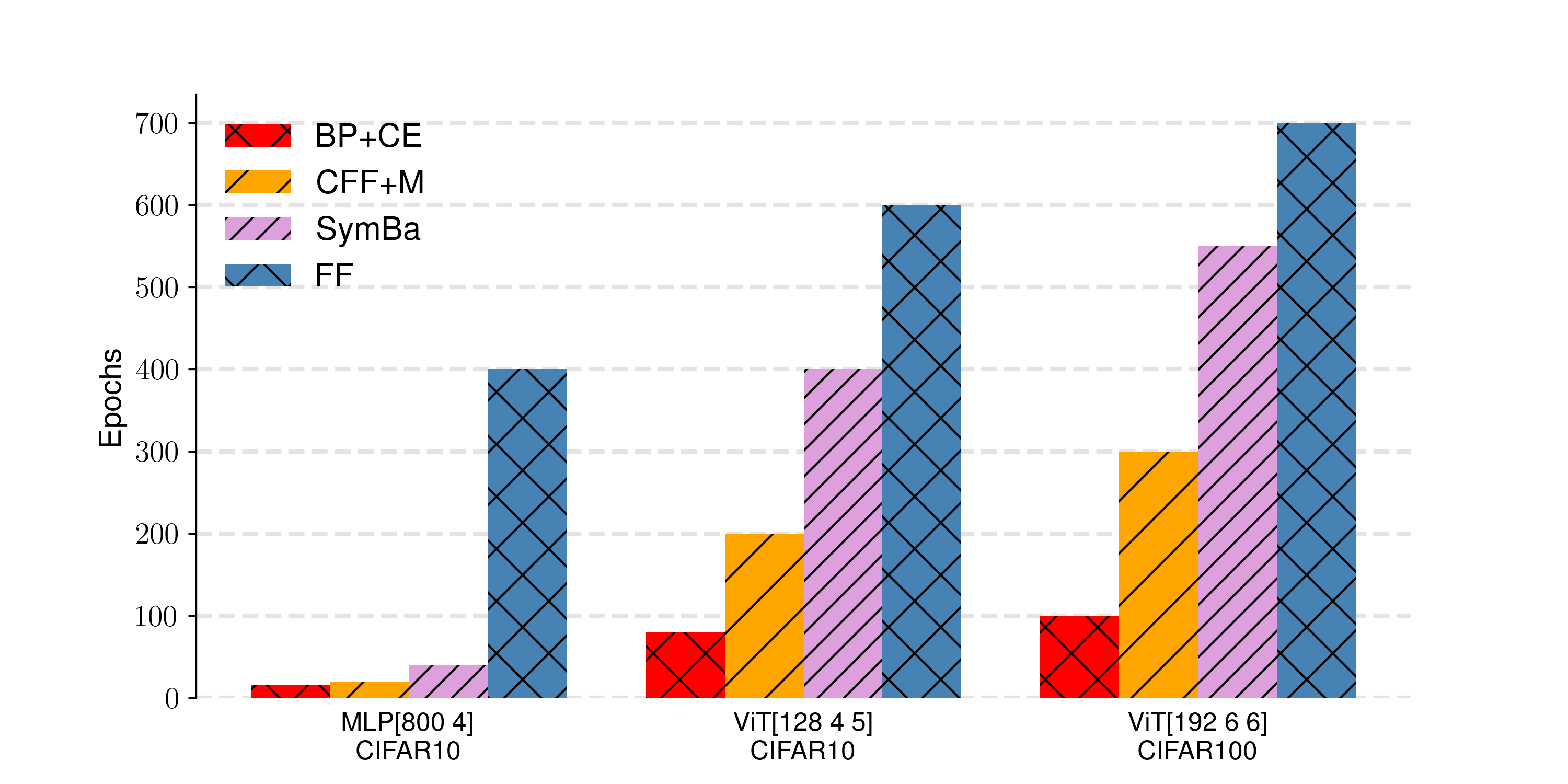}
   \caption{Comparison of the convergence speed of methods in three scenarios (model and dataset).}
   \label{fig:convergence}
\end{figure}

In Fig. \ref{fig:convergence}, the number of epochs required to reach the convergence point in various scenarios is presented. In all scenarios with FF-based Methods, the proposed algorithm demonstrated a faster convergence and it shows 3 to 20 times faster than baseline FF, although it is still slower than BP with CE loss function (BP+CE).

\subsection{How CFF Stacks Up Against Backpropagation}
\label{sec:cffvsbp}
Table \ref{tab:cffvsbp} shows the results of comparing the proposed algorithm with BP in terms of accuracy under different scenarios. We considered CE as loss function for BP since this loss function is often used in BP, and there are often doubts about the use of other loss functions in typical applications for better performance \cite{musgrave2020metric}. Thus, the encoder network along with the head is trained in an end-to-end with BP.

\begin{table}[H]
\footnotesize
\centering

\caption{Performance of proposed method (CFF+M) against backpropagation with cross entropy loss (BP+CE) on various datasets, augmentation (Aug.) and
models in terms of accuracy.\\~}
\begin{tabular}{c l l c c}
\toprule
Aug.& Dataset &Model & BP+CE & CFF+M (\scriptsize{\textit{ours}})\\
\midrule
\multirow{3}{*}{\rotatebox[origin=c]{90}{-}} & 
MNIST & 
MLP[500 3] &
\textbf{98.00}  &
97.30  \\
& 
MNIST &
MLP[800 4] &
\textbf{98.13}  & 
97.62  \\

&
CIFAR10&MLP[800 4]   &
\textbf{54.20}  &
\textbf{54.55} \\
\midrule
\multirow{6}{*}{\rotatebox[origin=c]{90}{Random Crop}} &
CIFAR10 &
ViT[128 4 5] &
76.92  &
\textbf{80.42} \\
&
CIFAR10 &
ViT[192 6 6]  &
76.68  &
\textbf{79.74} \\
&
CIFAR10 &
ViT[240 6 7]  &
77.04  &
\textbf{79.96} \\
& 
CIFAR100 &
ViT[192 6 6]  &
78.12  &
\textbf{80.39} \\
&
CIFAR100&
ViT[240 6 7] &
77.75  &
\textbf{78.44} \\
&
TImagenet &
ViT[240 6 7]  &
\textbf{74.01}  &
73.23 \\

\midrule
\multirow{6}{*}{\rotatebox[origin=c]{90}{\makecell{ Random Crop\\+ RandAug}}} &
CIFAR10 &
ViT[128 4 5]  &
\textbf{86.33}  &
81.73 \\
&
CIFAR10&
ViT[192 6 6] &
\textbf{86.95}  &
85.17 \\
&
CIFAR10 &
ViT[240 6 7] &
\textbf{87.62}  &
85.37 \\

&
CIFAR100&
ViT[192 6 6]  &
\textbf{84.13}  &
\textbf{84.28} \\
&
CIFAR100&
ViT[240 6 7]  &
82.88  &
\textbf{84.29 }\\
&
TImagenet&
ViT[240 6 7]  &
\textbf{78.95}  &
76.06 \\

\bottomrule

\end{tabular}
\label{tab:cffvsbp}
\end{table}

\begin{figure}[b]
    \centering
    \begin{subfigure}[b]{0.46\textwidth}
        \centering
        \includegraphics[width=\textwidth]{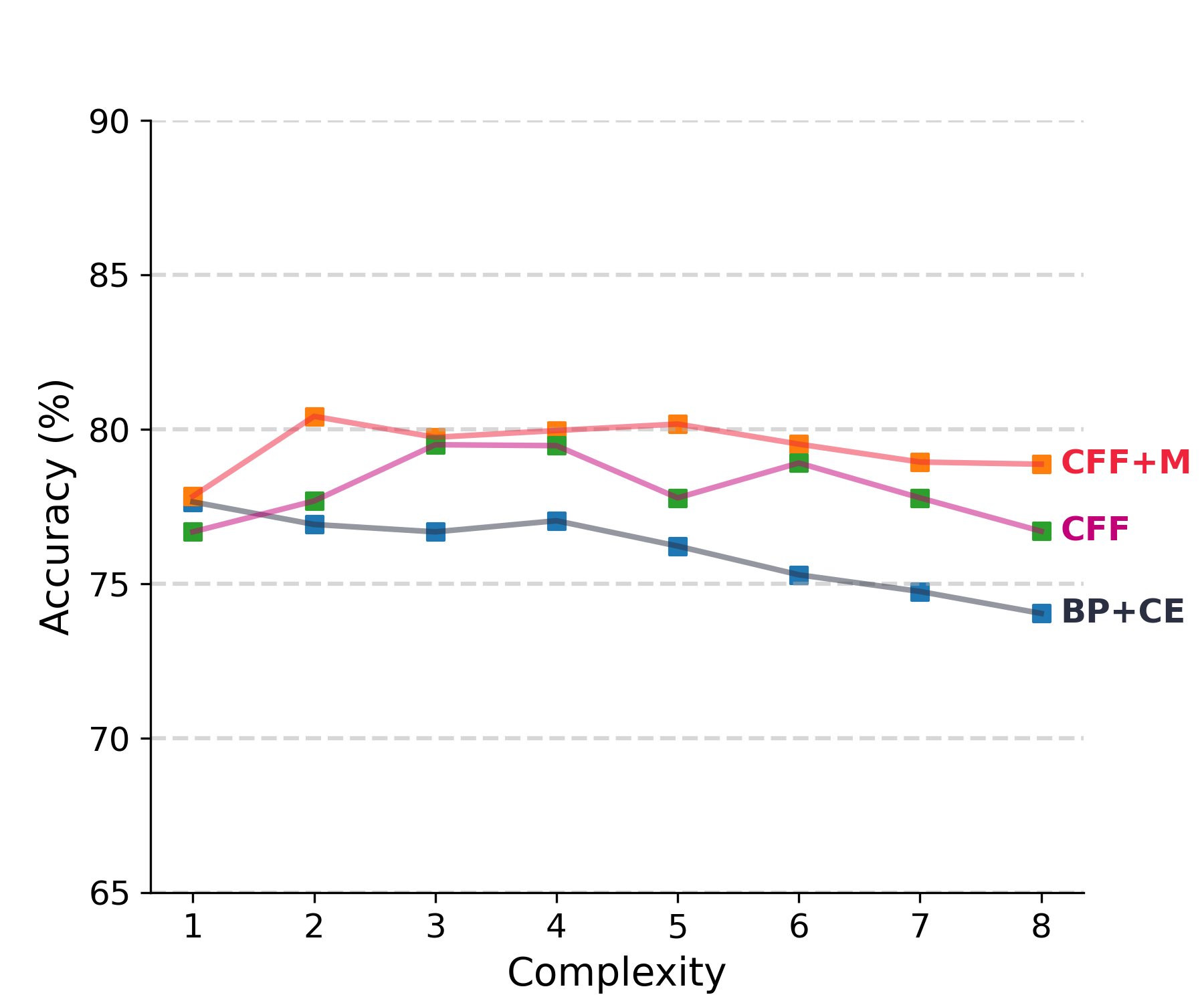}
        \caption{Augmentation: Random Crop}
        \label{fig:complexitya}
    \end{subfigure}
    \hfill
    \begin{subfigure}[b]{0.46\textwidth}
        \centering
        \includegraphics[width=\textwidth]{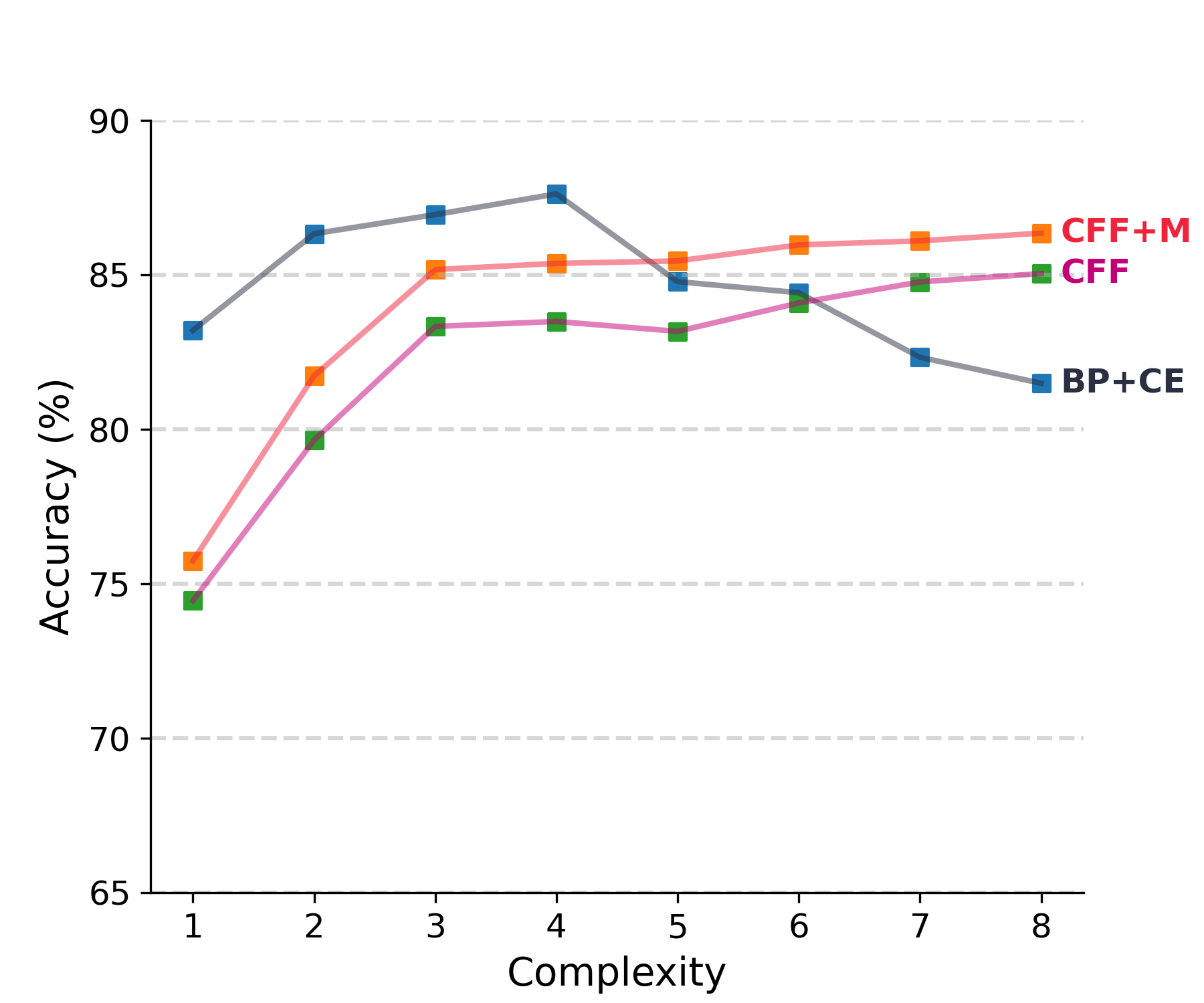}
        \caption{Augmentation: RandAug \cite{cubuk2020randaugment}}
        \label{fig:complexityb}
    \end{subfigure}
    \caption{Performance of proposed method (CFF \& CFF+M) against BP with cross entropy loss for CIFAR-10 on validation set: effect of model complexity. Model complexity is considered from 1 to 8 in the form of ViT[E H L] notation, where E = 64, 128, 192, 240, 280, 320, 360, 400, H = 4, 4, 6, 6, 7, 8, 9, 10, and L = 4, 5, 6, 7, 8, 9, 10, 11, respectively.}
    \label{fig:complexity}
\end{figure}
To compare the performance of the proposed algorithm with BP in ViT, three general scenarios were considered. In all three cases, horizontal flip augmentation is used. In the first scenario, which is related to MLP networks, no further data augmentation is used. In the second scenario, a well-known augmentation method called Random Crop is used, as in \cite{krizhevsky2012imagenet, he2016deep}. In this case, the proposed algorithm achieved better accuracy results across different datasets and ViT networks. In the third scenario, a highly effective augmentation method known as RandAug \cite{cubuk2020randaugment}, which itself includes several different augmentations, was used along with Random Crop. It is in a manner similar to approaches taken in state-of-the-art works on image classification problems \cite{beyer2022better, tan2021efficientnetv2, touvron2021training, liu2021swin}. In this scenario, BP performs better, although the proposed algorithm achieves a close performance.

Further analysis of Table \ref{tab:cffvsbp} reveals a specific pattern in comparing BP+CE with the proposed algorithm concerning model complexity and the type of augmentation used. It seems that CFF+M demonstrates higher robustness to overfitting compared to BP as model complexity increases. To illustrate this, Fig. \ref{fig:complexity} shows accuracy vs. model complexity under two augmentation scenarios. First, it is observed that using the proposed loss in the method (CFF+M) leads to better performance compared to using the original loss in the method (CFF) across all cases. Analysis of these plots shows that CFF+M tends to overfit at higher complexities compared to BP. This behavior can be attributed to an inherent form of regularization in FF-based method which layers have light collaboration. It can be particularly useful in applications where there is a higher risk of overfitting, such as when powerful augmentations like RandAug \cite{cubuk2020randaugment} are not employed or lack of training data. Furthermore, when using RandAug, which serves as the primary playing field for comparison, the proposed method can achieve competitive performance at higher complexities, which was obtained with BP in a lower model complexity scenario.

\subsection*{Robustness against inaccurate supervision and limited data problem}
In most practical applications, label noise is unavoidable, especially when dealing with datasets that require millions of annotations \cite{mahajan2018exploring}. Since previous experiments have shown that the proposed method is robust to overfitting compared to BP, it is expected to perform well in scenarios with limited data or data with noisy labels. Table \ref{tab:noise} shows the results of experiments in which a specific percentage of the CIFAR-10 data was either missing (limited data) or had noisy labels (inaccurate supervision). RandAug \cite{cubuk2020randaugment}  has been used to tilt the playing field in favor of BP. In the limited data scenario, a specific percentage of the training data was randomly removed. In the noisy label scenario, a specific percentage of the training data was incorrectly labeled. Although in the noise-free and data loss-free scenario, BP+CE performed better, in all other scenarios shown in the table, the proposed algorithm had less performance degradation and achieved higher accuracy. For example, in the scenario where 20\% of the labels were incorrectly assigned, BP experienced an accuracy drop of 8.85\%, whereas the proposed method (CFF+M) showed a 6.84\% decrease.

\begin{table}[H]
\footnotesize
    \centering
    \caption{Results on CIFAR-10 with ViT[192 6 6] for limited data and noisy labels scenarios. The numbers in parentheses indicate the reduction in accuracy (\%) compared to the scenario without label noise (N) or missing data (M).\\}
    \begin{tabular}{c c c c}
    \toprule
    M & N & BP+CE & CFF+M (\scriptsize{\textit{ours}})  \\
    \midrule
    
    0\%
    & 0\%
    & \textbf{86.95} 
    & 85.17 \\
    
    50\%
    & 0\%
    & 78.27 \scriptsize{(8.68\(\downarrow)\)}
    & \textbf{80.58 \scriptsize{(5.59\(\downarrow)\)}} \\
    
    80\%
    & 0\%
    & 69.19 \scriptsize{(17.76\(\downarrow)\)}
    & \textbf{73.26 \scriptsize{(12.91\(\downarrow)\)}} \\
    
    0\%
    & 20\%
    & 78.10 \scriptsize{(8.85\(\downarrow)\)} 
    & \textbf{78.33 \scriptsize{(6.84\(\downarrow)\)}} \\
    
    0\%
    & 40\%
    & 70.36 \scriptsize{(16.59\(\downarrow)\)}
    & \textbf{71.20 \scriptsize{(13.97\(\downarrow)\)}} \\
    \bottomrule
    \end{tabular}
    \label{tab:noise}
\end{table}

\subsection{Ablation Study}
\label{sec:ablative}
In this section, we demonstrate the impact of the algorithm's design elements through experiments.
\subsubsection*{Token vs Pooling}
As discussed in Sec. \ref{sec:cffvit}, there are two common approaches to classification using ViTs from a specific perspective. The first approach is to use class tokens, and the second is to use the mean output of layers for the loss function and final MLP layer. Table \ref{tab:tokenvspool} shows the experimental results comparing these two approaches for the CFF+M. The first approach is to feed the output of each class token into the contrastive loss function, while the second approach involves feeding the mean of the representations along each token element into the loss function. In the second approach, there is no class token. As can be seen from this table, the use of the second approach leads to better performance, which was also previously shown in \cite{beyer2022better} for BP.

\begin{table}[h]
\footnotesize
\centering
\caption{Ablation Study: Comparison of the using \textit{Token} versus \textit{Pooling} approach in the proposed
method in terms of accuracy (Acc.).\\~}
\label{tab:tokenvspool}
\begin{tabular}{c c c c}
\toprule
Model &  Dataset & Approach & Acc.\%\\
\midrule
\multirow{2}{*}{ViT[128 4 5]}
&\multirow{2}{*}{CIFAR10} & Token   & 76.63   \\
&                         & Pooling & \textbf{81.73}   \\
\midrule
\multirow{2}{*}{ViT[192 6 6]}
&\multirow{2}{*}{CIFAR100} & Token   & 79.00   \\
&                          & Pooling & \textbf{84.28}   \\
\bottomrule
\end{tabular}
\end{table}

\subsubsection*{One-forward vs Two-forward}
In SCL, two augmentations (views) of a batch of data are generated and passed through the network in two forward passes. This procedure is also carried out in CFF (the proposed method). In this way, a concatenation of the two obtained representations, $f = [f_1, f_2]$, is generated, where the corresponding representations $f_1$ and $f_2$ belong to the same images but with different views. Finally, the contrastive loss function (Eq. \ref{eq:supcon} or \ref{eq:msupcon}) calculates the loss by comparing each possible pair in $f$. As a result, the advantage of having $f_2$ is the availability of more views from the samples, and since different views will eventually be seen during training iterations, conceptually, it seems that $f_2$ might not be necessary. To test this hypothesis, an experiment was designed in which the proposed method works with only one forward pass ($f = [f_1]$). The results are presented in Table \ref{tab:onevstwo}. As can be seen from the table, one-forward approach can perform slightly weaker than two-forward approach, but it can be twice as fast in training per epoch, while it requires approximately the same number of epochs to converge. In this case, our proposed method can eliminate the need for two forward passes in the FF algorithm with a tolerable loss of  accuracy. Considering that the primary focus of the experiments is to achieve higher model accuracy and to adhere to SCL and FF, two-forward approach have been used in all previous experiments for the proposed method.

\begin{table}[t]
\footnotesize
\centering
\caption{Ablation Study: Comparison of \textit{One-Forward} (OF) approach versus \textit{Two-Forward} (TF) approach CFF+M in terms of accuracy (Acc.) and the time required for training per epoch (T.). RC: Random Crop, RA: RandAug \cite{cubuk2020randaugment}\\~}
\begin{tabular}{l c c c}
\toprule
Model  \tiny{Aug.} & Acc.(\%) \tiny{OF - TF} & T.(s) \tiny{OF/TF} & T.(s) \tiny{BP}\\
\midrule
\multicolumn{4}{c}{CIFAR10} \\
\midrule
ViT[128 4 5]  \tiny{RC} & -1.42 & \textbf{9}/18 &
8 \\
ViT[192 6 6]  \tiny{RC} & -0.04 & \textbf{14}/27 & 13  \\
ViT[192 6 6]  \tiny{RC+RA} & -1.92 & \textbf{23}/48 & 22 \\
\midrule
\multicolumn{4}{c}{CIFAR100} \\
\midrule
ViT[192 6 6]  \tiny{RC+RA} & -1.65 & \textbf{23}/48 & 22 \\
ViT[240 6 7]  \tiny{RC} & -1.08 & \textbf{19}/38 & 18 \\
\midrule
\multicolumn{4}{c}{Tiny Imagenet} \\
\midrule
ViT[240 6 7]  \tiny{RC+RA} & -2.20 & \textbf{56}/110 & 54  \\
\midrule
\end{tabular}
\label{tab:onevstwo}
\end{table}

\section{Discussion}
\label{sec:discussion}
\subsection{Layer Training Parallelization}
\label{sec:pip}
FF has made the layers independent of each other except for their input feed by placing the loss function after each layer and updating the weights of that layer accordingly. Although this minimizes the collaboration among layers \cite{lorberbom2024layer}, by continuing this approach, our proposed algorithm achieved close accuracy compared to BP on ViT (Sec. \ref{sec:cffvsbp}). However, the significant advantage of our proposed algorithm, inherited from FF, is the ability to create a pipeline of layers, enabling layer training parallelization. Since the weight updates of layers in this framework are independent of each other, while one layer is being updated, other layers can perform their weight updates in parallel. This can also be done during the forward pass of each layer. In contrast to BP, where operations are performed sequentially from the beginning to the end of the network during the forward pass and from the end to the beginning during the backward pass batch by batch \cite{rumelhart1986learning, hinton2022forward}.

\begin{figure}[t]
	\centering
	\includegraphics[width=1\linewidth]{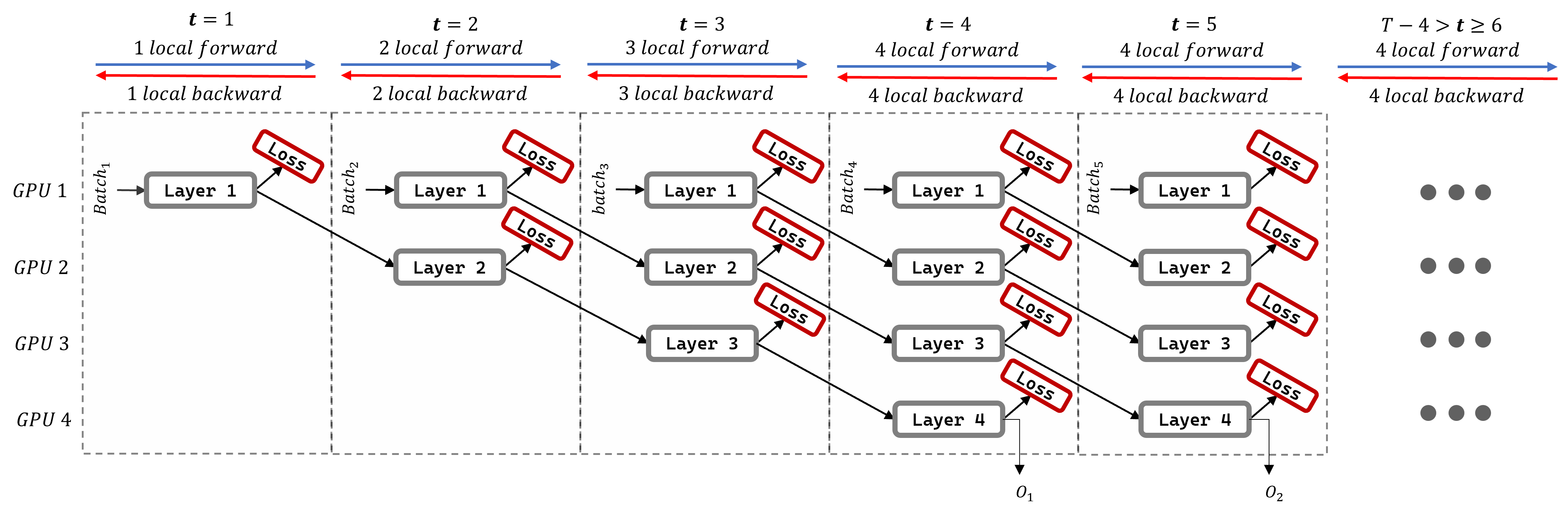}
	\caption{A schematic of the parallel processing approach in layers using a pipeline framework for the proposed method with the one forward approach. $O_b$ is the output of the encoder belonging to $Batch_b$.}
	\label{fig:pip}
\end{figure}

Fig. \ref{fig:pip} illustrates the proposed pipeline. In this figure, an example with a 4-layer network using 4 GPUs is considered. For simplicity, the one-forward approach is used (Sec. \ref{sec:ablative}). The entire figure pertains to one epoch, where each step is an iteration along the batches. At $t=1$, the first batch enters the first layer. The local forward and local backward processes in this layer are performed by GPU 1. The output of this layer is transferred to the second layer at $t=2$. Since the first layer and GPU 1 are idle, they can take in the second batch, while at the same time, GPU 2 processes the second layer. As a result, in addition to preparing the output of the second batch in the first layer, the output of the second layer belonging to the first batch is also ready; And following the same procedure up to $t=4$. At $t=4$, the layers are processing representatives from layers 1 to 4 for batches 4 to 1, respectively and there are effectively 4 batches within the network. In this case after $t\geq4$, the output for a batch is ready at each step from the viewpoint of the whole network. Therefore, if the total number of batches is $N$ and the total number of required steps is $T$, the first batch requires 4 steps to pass through the entire network, while the output for the second batch is ready at $t=5$, and similarly, the output for subsequent batches is ready at each step. Hence, $T=4+(N-1)$. For BP, each batch requires 4 steps, so $T=4N$, which means that this parallelization, approximately speeds up training by $4$ in each epoch. Fortunately, each transformer layer shown in Fig. \ref{fig:overvit} has the same structure. Consequently, it is expected that processing time at each layer with a fixed batch size will be equal, which reinforces the assumptions made. However, a challenge that is not considered in this example is the IO processing time. Since batches in typical training methods are read sequentially from storage, this can create a processing time bottleneck.

When multiple GPUs are available, different types of parallelization are also possible for BP. One common method is to replicate the entire model to each GPU and split the batches across the GPUs \cite{paszke2019pytorch}. The forward pass is performed independently on each GPU, and the backward pass is carried out based on the averaged gradients obtained. In this case, increasing the batch size will speed up the training process. However, we believe that this approach in BP cannot achieve the same efficiency as our proposed pipeline, where instead of placing the entire model on each GPU, each layer is distributed across GPUs.

In an experiment utilizing two Tesla T4 GPUs and applying the proposed algorithm with the one forward approach on ViT[128 4 2] (2 layers), CFF achieves approximately 1.8 times faster training speed per epoch compared to BP. To achieve parallelization in BP, torch.nn.DataParallel \cite{paszke2019pytorch} is used, which is a straightforward and commonly used approach when having multiple GPUs for BP. In the experiment to avoid the I/O bottleneck challenge, all data was kept in RAM. Given the implementation challenges and the unavailability of a high number of GPUs, we have left the focus on the pipeline feature of the proposed algorithm to future works.

\subsection{Encounter of FF-based Methods with Potential Layer's Inductive Bias}

FF was introduced for use in MLP networks \cite{hinton2022forward}. One of the goals of this work is to evaluate the performance of FF-based methods on a more practical network. The foundation of CNNs is formed by convolutional operators. The standard blocks in these networks typically include two convolutional layers with nonlinear activation functions, along with a skip connection, pooling, etc \cite{tan2021efficientnetv2, liu2022convnet}. Compared to the ViT block (Encoder Layer) or even an MLP  block, such a structure does not significantly alter the input. This phenomenon can be attributed to the strong inductive bias present in convolutional blocks which is due to the local operations \cite{dosovitskiy2020image}. Indeed, one of the main reasons for the greater depth of CNNs compared to ViTs is the relatively lower power of a single CNN block compared to a transformer block. The inductive bias in CNNs offers many advantages, such as requiring significantly fewer data compared to ViTs \cite{dosovitskiy2020image}. However, this characteristic is not beneficial in algorithms based on FF, including our proposed algorithm. These algorithms operate on the similarity and dissimilarity of representations after each layer, and convolutional blocks, especially in the initial layers, cannot produce similar representations for samples of the same class until the final layers. We believe that this challenge, considering the high number of layers in CNNs, leads to poor performance of the proposed algorithm in these networks, and this may be the reason why in \cite{papachristodoulou2024convolutional} that is a FF-based method, the foundational convolutional structure was modified. In the same way, we believe that applying any form of inductive bias, such as local attention mechanisms on ViT \cite{yuan2021tokens}, will degrade the performance of FF-based methods. On the other hand, reducing inductive bias increases the risk of overfitting. In contrast, BP considers the entire network, which provides more flexibility for adjusting inductive biases, such as tuning the number of layers.

\section{Conclusion}\label{sec:conclusion}
FF \cite{hinton2022forward} is a novel method for training neural networks, specifically designed for image classification. This algorithm calculates a loss function after each layer, making the training of the layers independent of each other, except for preparing their inputs. This approach is different from the traditional BP algorithm, which employs the CE loss function (but is not limited to this loss) at the end of the network. From this point of view, FF is more plausible to brain function, where it has been shown that BP is not \cite{lillicrap2020backpropagation}.

FF uses two forward passes: one with the correct input label and another with an incorrect label. Layers are trained by comparing the two resulting representations. This comparison between the two representations can also be found in another well-known field named Contrastive Learning \cite{hadsell2006dimensionality, sohn2016improved, chen2020simple, khosla2020supervised}. Therefore, we used insights from it to modify FF. We replaced the original FF loss function with a supervised contrastive loss and adapted the input sample feeding method to align with SCL \cite{khosla2020supervised}. Thus, the proposed method can be viewed from two points of view: as SCL where the loss function is placed after each layer instead of at the end (similar to FF), or as FF with contrastive loss.

FF was initially introduced based on MLP networks. We believe that although FF and the proposed method can be implemented on a wide range of networks, it may not work well in a large part of them, especially CNNs. However, ViT \cite{dosovitskiy2020image} with its more complex layers, could be better suited to this algorithm. In this work, after presenting the proposed algorithm, the application of FF and the proposed algorithm was extended to ViT without any changes to its network architecture.

Experiments demonstrated that the proposed method outperformed the baseline FF in terms of accuracy, inference time (against FF without head) and convergence speed. Additionally, to better integrate contrastive loss into FF, we introduced the Marginal Contrastive Loss by slightly modifying the original contrastive loss. Experiments showed that this led to better results. Although our primary goal was to improve FF, another goal was to develop an algorithm that is able to compete with BP. Additional experiments show that the proposed algorithm can achieve relatively close and competitive performance with BP using the CE loss function on ViT.

The proposed algorithm has some disadvantages compared to BP including the limited choice of loss functions, relatively
complex implementation, and restriction to classification problem. Extending the method to complex vision tasks
like object tracking/detection and image segmentation, evaluating it on large-scale models and bigger datasets, and
examining its biological plausibility are suggested as future work.
\renewcommand{\theequation}{A.\arabic{equation}}
\setcounter{equation}{0}
\appendix
\section{Appendix}

\subsection{Gradient Calculation}\label{sec:gradient}
Recall that $f_k$ is the representation obtained from an encoder layer for the $k$-th sample in the batch, $A(k) = P(k) \cup N(k)$ and $P(k) = R(k) \cup R^c(k)$. To compute the gradient of Marginal Contrastive Loss with respect to $f_k$ (for an arbitrary $k$), from Sec. \ref{sec:msupcon} recall that:

\begin{equation}
L_i = \left(- \frac{1}{|P(i)|} \sum_{p \in P(i)} Q(f_i,f_p)\right) + \log \left( \sum_{p' \in P(i)} \exp(Q(f_i,f_{p'})) + \sum_{n \in N(i)} \exp(f_i\cdot f_n) \right)
\label{eq:li}
\end{equation}

one of the following scenarios will occur:

\paragraph{1. \( f_k \) is the anchor}
~\\ Consider Eq. \ref{eq:li} with $k$ substituted for $i$. In this way:

\begin{equation}
\frac{\partial L_k}{\partial f_k} = - \frac{1}{|P(k)|} \sum_{p \in P(k)} \frac{\partial Q(f_k,f_p)}{\partial f_k} + \frac{\sum_{p' \in P(k)} \exp(Q(f_k,f_{p'})) \frac{\partial Q(f_k,f_{p'})}{\partial f_k} + \sum_{n \in N(k)} \exp(f_k\cdot f_n) f_n}{\sum_{p' \in P(k)} \exp(Q(f_k,f_{p'})) + \sum_{n \in N(k)} \exp(f_k\cdot f_n)}
\end{equation}

According to Eq. \ref{eq:msupcon}:

\begin{equation}
Q(f_a,f_b) = 
\begin{cases} 
    1 & a \in R(b) \\
    f_a\cdot f_b + m & a \in R^c(b)
\end{cases}
\label{eq:case}
\end{equation}

As a result, the gradient is obtained as shown in Eq. \ref{eq:lossg}.
\paragraph{2. $f_k$ is not the anchor}
~\\There is an $i \neq k$ that serves as index of the anchor. Considering Eq. \ref{eq:li}, depending on whether \( k \in P(i) \) or \( k \in N(i) \):

\paragraph{2.1. $k \in P(i)$}
\begin{equation}
\frac{\partial L_{i,k \in P(i)}}{\partial f_k} = - \frac{1}{|P(i)|} \frac{\partial Q(f_i,f_k)}{\partial f_k} + \frac{\exp(Q(f_i,f_k)) \frac{\partial Q(f_i,f_k)}{\partial f_k}}{\sum_{p' \in P(i)} \exp(Q(f_i,f_{p'})) + \sum_{n \in N(i)} \exp(f_i\cdot f_n)}
\label{eq:lossgnotanchor}
\end{equation}

Depending on whether \( k \in R(i) \) or \( k \in R^c(i) \):
\paragraph{2.1.1. \( k \in R(i) \)}
~\\according to Eq. \ref{eq:lossgnotanchor}:
\begin{equation}
\frac{\partial L_{i,k \in P(i), k \in R(i)}}{\partial f_k} = 0
\label{eq:lossg2}
\end{equation}
\paragraph{2.1.2. \( k \in R^c(i) \)}
~\\according to Eq. \ref{eq:lossgnotanchor}:\begin{equation}
\frac{\partial L_{i,k \in P(i), k \in R^c(i)}}{\partial f_k} = \frac{1}{|P(i)|} f_i + \frac{\exp(f_i\cdot f_k + m) f_i}{|R(i)| + \sum_{r' \in R^c(i)} \exp(f_i\cdot f_{r'} + m) + \sum_{n \in N(i)} \exp(f_i\cdot f_n)}
\label{eq:lossg3}
\end{equation}
\paragraph{2.2. \( k \in N(i) \)}
~\\According to Eq. \ref{eq:li}:

\begin{equation}
\frac{\partial L_{i,k \in N(i)}}{\partial f_k} = \frac{\exp(f_i\cdot f_k) f_i}{\sum_{p' \in P(i)} \exp(Q(f_i, f_{p'})) + \sum_{n \in N(i)} \exp(f_k\cdot f_n)}
\label{eq:lossg4}
\end{equation}

\subsection{Datasets}
\label{sec:Adataset}

In the experiments, we used four benchmark and public datasets. These datasets are as follows: MNIST \cite{lecun1998gradient}, which consists of 60,000 training images and 10,000 test images in grayscale, with a size of 28x28 pixels. CIFAR-10 \cite{krizhevsky2009learning}, comprising 50,000 training images and 10,000 test images across 10 classes such as airplanes, cars, dogs, cats, etc. These images have three RGB color channels and a size of 32x32 pixels. CIFAR-100 \cite{krizhevsky2009learning}, which is similar in quantity and size to CIFAR-10 but with 100 classes. Tiny ImageNet \cite{le2015tiny}, a subset of the well-known ImageNet dataset \cite{deng2009imagenet}, contains 100,000 training samples and 10,000 test samples. The images have three RGB color channels with a size of 64x64 pixels. The batch size was set to 512 in all cases. The patch size on ViT for CIFAR-10 and CIFAR-100 was set to 4x4, and for the Tiny ImageNet was set to 8x8. In both cases, this results in 64 patches being generated.

\bibliographystyle{unsrt}
\bibliography{references}  

\end{document}